\documentclass[11pt, letterpaper, copyright, logo]{googledeepmind}

\usepackage{multirow}
\usepackage[authoryear, sort&compress, round]{natbib}
\usepackage{tabularx}
\usepackage{enumitem}
\usepackage{lipsum}
\usepackage{amsmath}
\usepackage{pstricks, pst-node}
\usepackage{verbatim}
\usepackage{multirow}
\usepackage{multicol}
\usepackage{longtable}
\usepackage{array}
\usepackage{listings, listings-rust}
\usepackage{booktabs}
\usepackage{enumitem}
\usepackage{xspace}
\usepackage{bm}
\usepackage{bbm}
\usepackage{mathtools}
\usepackage{soul}
\usepackage{graphicx}
\usepackage[most, breakable, skins]{tcolorbox}
\usepackage{subcaption}
\usepackage{amssymb}
\usepackage{colortbl}
\usepackage{csquotes}
\usepackage[inkscapeformat=png]{svg}
\usepackage{tabularx,ragged2e}
\usepackage{makecell}
\usepackage{placeins}
\usepackage[dvipsnames]{xcolor}
\usepackage{hyperref}
\usepackage{pifont}
\usepackage[capitalise]{cleveref}
\usepackage{threeparttable}

\newcolumntype{L}[1]{>{\raggedright\let\newline\\\arraybackslash\hspace{0pt}}m{#1}}
\newcolumntype{C}[1]{>{\centering}m{#1}}

\newcolumntype{R}[1]{>{\raggedleft\let\newline\\\arraybackslash\hspace{0pt}}m{#1}}

\definecolor{ao}{rgb}{0.0, 0.0, 1.0}

\newcommand\vcent[1]{\vcenter{\hbox{#1}}}
\newcommand\loudspeaker[1][3]{\ensuremath{\vcent{\rule{.6ex}{.6ex}}\kern-.5ex%
  \vcent{\scalebox{.6}[1]{\rotatebox[origin=center]{90}{$\blacktriangle$}}}%
  \ifnum#1>0\relax\kern.05ex\vcent{\scalebox{.4}{\ttfamily)}}%
  \ifnum#1>1\relax\kern-.4ex\vcent{\scalebox{.56}{\ttfamily)}}%
  \ifnum#1>2\relax\kern-.55ex\vcent{\scalebox{.7}{\ttfamily)}}%
  \fi\fi\fi}%
}

\newcommand{\tick}{\ding{51}}

\let\cite\citep

\title{MedGemma Technical Report}

\author[]{Google Research and Google DeepMind \footnote{\small{ See Contributions and Acknowledgments section for full author list. \\\hspace*{2.7em}\small{Corresponding authors: \{linyan, dangolden, asellerg\}@google.com}.}}}

\begin{document}

\begin{abstract}
Artificial intelligence (AI) has significant potential in healthcare applications, but its training and deployment are challenging due to healthcare's diverse data, complex spectrum of possible tasks, and the need to preserve privacy. Foundation models that perform well on various medical tasks and require less task-specific tuning data are critical to accelerating the development of AI for healthcare applications. In this technical report, we introduce MedGemma, a new collection of medical vision–language foundation models based on Gemma 3 4B and 27B. MedGemma demonstrates advanced medical understanding and reasoning on images and text, significantly exceeding the performance of similar-sized generative models and approaching the performance of task-specific models, while maintaining the general capabilities of the Gemma 3 base models. For out-of-distribution tasks, MedGemma achieves 2.6-10\% improvements on medical multimodal question answering, 15.5-18.1\% improvements on chest X-ray finding classification, and 10.8\% improvement on agentic evaluations compared to the base models. Fine-tuning MedGemma further improves performance in subdomains, reducing errors in electronic health record information retrieval by 50\% and reaching comparable performance to existing specialized state-of-the-art methods for pneumothorax classification and histopathology patch type classification. We additionally introduce MedSigLIP, a medically-tuned vision encoder derived from SigLIP. MedSigLIP powers the visual understanding capabilities of MedGemma and, as an encoder, it achieves performance comparable to or better than specialized medical image encoders. Taken together, the MedGemma collection provides a strong foundation of medical image and text capabilities, with potential to significantly accelerate medical research and development of downstream applications. More details about the MedGemma collection, including tutorials and instructions for downloading the model weights, can be found at \url{https://goo.gle/medgemma}.
  
\end{abstract}

\maketitle

\clearpage
\section{Introduction}
\label{sec:intro}
The landscape of modern healthcare is characterized by the generation and use of an unprecedented volume and diversity of data. Diagnosis, treatment, and monitoring rely on synthesizing information from disparate sources and specialties. Recently developed large multimodal models (LMMs), trained on massive and diverse datasets, exhibit remarkable capabilities in detecting complex patterns, generating coherent text, and processing visual information \cite{alayrac2022flamingo, chen2022pali, achiam2023gpt, openai2023gpt4v, touvron2023llama, liu2023improved,liu2024visual}. These capabilities mark a potential paradigm shift in assisting with current workflows and extracting novel insights.

While general-purpose (non-medically tuned) LMMs demonstrate impressively broad abilities, generic models can lack nuanced medical understanding and the ability to interpret and reason about medical data in a robust way \cite{singhal2023large,singhal2023towards,toma2023clinical,han2023medalpaca,labrak2024biomistral,tu2024towards,yang2024advancing}. Recognizing this gap, we created MedGemma, a new suite of open, medically-tuned, vision-language foundation models. These models represent the latest addition to the Health AI Developer Foundations~\cite{kiraly2024health} collection. Built upon the robust architecture of Gemma 3 \cite{team2025gemma}, the MedGemma models are designed to interpret and reason about medical images and text while retaining the strong general-purpose capabilities present in Gemma 3.

\begin{figure}[htp]
    \centering
    \includegraphics[width=0.87\linewidth]{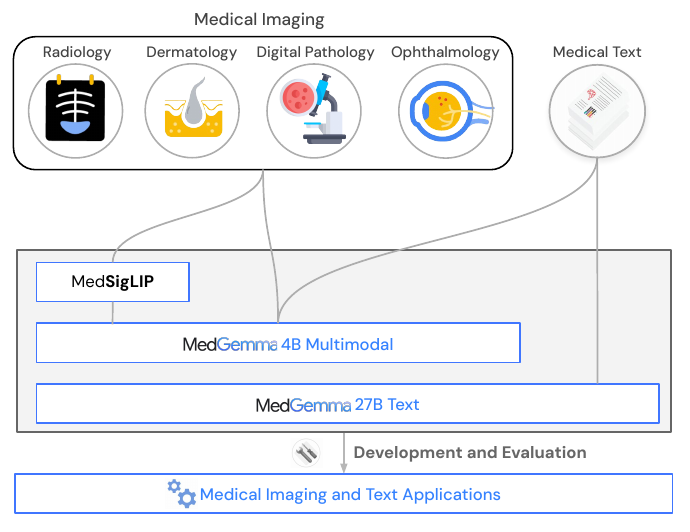}
    \caption{\small{\textbf{Overview of the MedGemma model collection} featuring the MedSigLIP image encoder, MedGemma 4B Multimodal and MedGemma 27B Text}}
    \label{fig:medgemma_overview}
\end{figure} 

In this report, we focus on two MedGemma models: a 4B variant that can accept text, images, or both as input, and a 27B variant that is optimized for text-only inputs. Both models output text. MedGemma 4B demonstrates strong performance on Vision Question Answering (VQA) benchmarks compared to prior SOTA models like Med-Gemini \cite{saab2024capabilities, yang2024advancing} despite being considerably smaller. Both MedGemma 4B and 27B are highly competitive on challenging text-only medical benchmark tasks, including MedQA \cite{medqa}, MedMCQA \cite{medmcqa}, PubMedQA \cite{pubmedqa}, MMLU Med \cite{mmlu}, AfriMed-QA \cite{afrimed}, and AgentClinic \cite{schmidgall2024agentclinic} when compared against other open models of similar scale. In addition to these strong out-of-the-box capabilities, we show how performance can be further improved by fine-tuning MedGemma on subdomains like chest X-ray reporting, histopathology classification, and electronic health record information retrieval.

An additional MedGemma variant, a multimodal version of MedGemma 27B, was also developed and is being released along with the other models. More thorough evaluation of this multimodal 27B variant is ongoing and preliminary results can be found in Appendix Section~\ref{sec:27bmm}. Unless otherwise noted in this report, evaluations that reference ``MedGemma 27B'' refer to the text-only variant of MedGemma 27B.

In addition to the MedGemma models, we describe the standalone MedSigLIP \mbox{400M-parameter} medical image encoder. MedSigLIP is based on \mbox{SigLIP-400M} \cite{zhai2023sigmoid} and is the same encoder that powers MedGemma's image interpretation capabilities. When used on its own, MedSigLIP enables data-efficient and zero-shot image classification and retrieval, with performance comparable to or exceeding specialized image encoders.

A high level overview of the released models is shown in \cref{fig:medgemma_overview}. More details about the MedGemma collection, including tutorials and links to download all of the above models, can be found at \url{https://goo.gle/medgemma}.

\section{Methods}
\label{sec:methods}
\subsection{Datasets}

For general purpose data replay during pretraining, original data mixtures from SigLIP \cite{zhai2023sigmoid} and Gemma 3 \cite{team2025gemma} were leveraged. The medical training and evaluation datasets largely followed the datasets in Med-Gemini \cite{yang2024advancing}. In this section, we outline the specific changes or differences in datasets relative to Med-Gemini.

\subsubsection{Training datasets}

\begin{table}[tbh]
\centering
\caption{\small{\textbf{Overview of MedGemma training datasets.}}}
\footnotesize
\label{tab:training-dataset-info}
\resizebox{\textwidth}{!}{
\renewcommand{\arraystretch}{1.25}
\begin{tabular}{l|l|r|l|p{2.6in}}
\Xhline{2.5\arrayrulewidth}
\textbf{Modality} & \textbf{Dataset}   & \textbf{No. examples} & \textbf{Training stages} & \textbf{Description} \\
\Xhline{2\arrayrulewidth}
\multirow{8}{*}{Text-only} & MedQA                     &  9,275 & Distill, RL & USMLE style exam questions \\
                           & MedMCQA                     & 182,806 & Distill, RL & Indian medical entrance exam questions \\
                           & AfriMed-QA                     & 1,003 & Distill, RL & Pan-African English multi-specialty QA \\
                           & MedExpQA
                           & 434 & Distill, RL & Spanish medical residency exam questions \\
                           & PubMedQA
                           & 1,000 & Distill, RL & Biomedical QA compiled from PubMed abstracts \\
                           & LiveQA
                           & 634 & Distill, RL & Consumer health questions from United States National Library of Medicine (NLM) \\
                           & HealthSearchQA
                           & 3,375 & Distill, RL & Common consumer medical questions from search engines \\
                           & Synthetic
                           & 200,000 & Distill & Generated from large IT teacher \\
\hline 
\multirow{5}{*}{Radiology}  & SLAKE              & 450       & Vision, PT, RL & Captions generated from QA pairs \\
                                 & VQA-Rad         & 1,391   & RL & Radiology image QA pairs \\
                                 & MIMIC-CXR            & 231,483  & Vision, PT, RL & Chest X-ray images \& free-form reports \\
                                 & Digital Knee X-ray       & 1,469  & Vision, PT   & Knee X-ray images \& labels \\
                                 & CT-US1                  & 59,979   & Vision, PT  & 2D CT slices \& free-form reports \\ 
                                 & MRI-US1              & 47,622     & Vision, PT   & 2D MRI slices \& free-form reports \\ 
                                    
\hline 
\multirow{1}{*}{Histopathology}   & Internal histopathology          & 32,550,599  & Vision, PT, RL & Histopathology image patches, caption pairs\\
\hline                              
\multirow{1}{*}{Dermatology} & PAD-UFES-20                 & 2,047   & Vision, PT  & Skin lesion images \& labels \\
                             & Internal dermatology        & 51,049  & Vision, PT, RL   & Skin lesion images \& labels \\
\hline 
\multirow{1}{*}{Ophthalmology} & EyePACS                    & 199,258   & Vision, PT, RL & Fundus images \& labels \\
\hline 
\multirow{1}{*}{General Medical} & PMC                     & 41,853 & Vision, PT & Single panel medical images \& caption pairs\\
\hline 
\Xhline{2.5\arrayrulewidth}
\end{tabular}
}
\begin{tablenotes}
  \scriptsize
  \item Vision: Vision encoder enhancement, PT: Pretraining, Distill: Distillation, RL: Reinforcement learning.
\end{tablenotes}
\end{table}

\paragraph{Text-only datasets:} For text datasets, we sampled responses and logits from a large IT (instruction-tuned) teacher using the train splits of multiple medical QA datasets, including MedQA~\cite{medqa}, MedMCQA~\cite{medmcqa}, PubMedQA~\cite{pubmedqa}, MedExpQA~\cite{medexpqa}, AfriMed-QA~\cite{afrimed}, HealthSearchQA~\cite{healthsearchqa}, and LiveQA~\cite{liveqa}. We also sampled responses and logits for approximately 200,000 synthetic medical questions generated by asking the same large IT teacher to generate a new question using 5 randomly sampled questions from the above datasets as examples.

\paragraph{Multimodal datasets:} Relative to Med-Gemini, the multimodal capabilities of MedGemma are currently focused on 2D medical images (e.g. X-ray, 2D slices from CT/MRI); 3D volumes and genomic datasets described in \citet{yang2024advancing} were not included. Additionally, we and others have identified potential data quality issues in PathVQA and MedVQA. Thus, we removed them from the training dataset. We did not include PAD-UFES-20 in the post-training dataset since it focuses on 6-class classification of very specific lesion types, which is not in line with the goal of more general purpose dermatology capabilities and use cases. For the PMC-OA component of the training data, we only included the single panel medical images from PMC-OA for better data quality. Relative to Med-Gemini we also introduced a larger internal collection for ophthalmology (184,852 more retinal fundus images), dermatology (51,049 more dermatology images with 210 different skin conditions), histopathology (a total of $\sim$32.5 million patch-text pairs), and radiology data (54,573 more CT 2D slices, 47,622 more MRI 2D slices). The additional CT and MRI slices utilized for training were curated based on mention of a specific slice associated with abnormal findings in the radiology report.

\subsubsection{Data Preprocessing}

Our data preparation followed~\citet{yang2024advancing} closely. Image padding and resizing algorithms remain the same, but because the vision encoder is different in Gemma 3, our images were resized to 896$\times$896 instead of 768$\times$768. Following Gemma 3, we use the SentencePiece tokenizer with 262,000 entries. Additionally, for CT images, we preselected three windows and converted them into the RGB color channels of the input image to highlight (1) bone and lung, window-width: 2250, window-level: -100; (2) soft tissue, window-width: 350, window-level: 40; (3) brain, window-width: 80, window-level: 40.

\subsection{Modeling Methodology}

\subsubsection{Modeling Architecture and Training Infrastructure}
The MedGemma model architecture follows Gemma 3 \cite{team2025gemma} and is compatible with all existing Gemma infrastructure. The vision encoder for Gemma 3 is the 400M variant of the
SigLIP encoder \cite{zhai2023sigmoid} and is shared across the different Gemma language model sizes (4B, 27B). The input image resolution is 896$\times$896 with pixel values normalized to [-1, 1]. The language model component also follows Gemma 3, featuring arbitrary image-text interleaving and long context (128k). Similar to Gemma 3, MedGemma was trained on TPUv4, TPUv5e, and TPUv5p, leveraged pre-computed visual tokens for memory saving, and used data and model shardings for multi-pod training.

\subsubsection{Model Training}

The MedGemma 4B multimodal model utilized all of the following steps while the text-only version of MedGemma 27B leveraged the post-training stage alone.

\paragraph{Vision Encoder Enhancement for MedGemma:} To improve the vision encoder’s capability of encoding and distinguishing subtle differences in medical images, we fine-tuned the vision encoder in Gemma 3 (SigLiP-400M) using over 33M medical image-text pairs (635k from various medical modalities and 32.6M histopathology patches) as listed in Table~\ref{tab:training-dataset-info}. To retain SigLIP’s existing performance, its original training data (e.g., WebLI) were retained and medical data was mixed with 2\% weight into the training. While the Gemma 3 vision encoder works with 896$\times$896 resolution, we found that many medical vision tasks worked reasonably well at 448$\times$448 resolution (Table ~\ref{tab:med_siglip_results}). Thus, while the MedGemma 4B image encoder is based on 896$\times$896 resolution for compatibility and consistency with Gemma 3, the released MedSigLIP model is based on 448$\times$448 resolution for more efficient experimentation and adaptation by the community. The 448$\times$448 encoder shares the same model weights as the 896$\times$896 encoder with the only difference being down-sampled positional embeddings to work with fewer input patches from the lower resolution.

\paragraph{Multimodal Decoder Pretraining:} After the vision encoder enhancement, the Gemma language model needed to be re-adapted for this new vision encoder, not only for the medical data but also for the general image domain to preserve the visual-language reasoning capabilities. This goal was achieved in the pretraining stage incorporating both the text and interleaved imaging data from the original mixture and the newly introduced medical domain image-text paired data. Notably, we did not introduce further medical text-only data in this step, as the original Gemma 3 mixture was already general-purpose and large-scale. To reduce compute requirements, we continued our pretraining on top of the original Gemma 3 pretrained checkpoints, mixed our medical image data (Table~\ref{tab:training-dataset-info}) with 10\% weight, trained for approximately 5 epochs on the medical mixture given the mixing ratio, and picked the checkpoint based on the validation set performance on chest X-ray report generation, and radiology, dermatology, and ophthalmology visual question answering.

\paragraph{Post-training:} The knowledge acquired from pretraining needs to be surfaced as capabilities in the post-training stage. There are two primary post-training components as previously outlined for Gemma 3. The recipes for distillation and reinforcement learning (RL) were the same as in Gemma 3 development with the following additions: (1) Distillation: addition of medical text data during this component to enable further learning in these domains from a large instruction-tuned (IT) teacher.
(2) Reinforcement learning:  Medical imaging data with paired text was utilized in the RL stage of post-training. For multimodal training, we found that RL enables better generalization compared to supervised fine-tuning, so all multimodal post-training was performed via RL.

\section{MedGemma Evaluations}
\label{sec:eval_medgemma}
MedGemma was evaluated and compared with other models on five types of medical tasks: text question-answering, image classification, visual question answering, chest X-ray (CXR) report generation, and agentic behavior. We additionally validated MedGemma on several general purpose (non-medical) benchmarks.  A high-level overview of tasks and datasets is provided in Table~\ref{tab:evals_data} and additional details on each task are below.

\begin{table}[htb]
\centering
\footnotesize
\caption{\small{\textbf{Overview of MedGemma evaluation datasets.}}}
\label{tab:evals_data}
\renewcommand{\arraystretch}{1.15}
\begin{tabular}{@{}l|l|l|r|c}
\Xhline{2.5\arrayrulewidth}
\textbf{Task}  & \textbf{Dataset}& \textbf{Modality}  & \textbf{No. Examples} & \textbf{OOD\textsuperscript{$\dagger$}}                 \\
\Xhline{2\arrayrulewidth}
\multirow{6}{*}{\begin{tabular}[c]{@{}l@{}}Medical text question-answering\end{tabular}}
& MedQA            & Text & 1,273   & -     \\ 
& MedMCQA          & Text & 4,183   & -     \\ 
& PubMed QA        & Text & 500   & -     \\ 
& MMLU Med         & Text & 3,685   & -     \\ 
& MedXpertQA       & Text & 2,450   & \tick \\ 
& AfriMed-QA (MCQ) & Text & 25   & -     \\ 
\hline
\multirow{6}{*}{\begin{tabular}[c]{@{}l@{}}Medical image classification\end{tabular}}
& MIMIC-CXR (Med-Gemini test set) & Radiology   & 1532   & -      \\ 
& MIMIC-CXR (MAIRA test set)      &  Radiology  & 2461   & -       \\
& ChestX-ray14 (CXR14)            & Radiology      & 1,962   & \tick      \\ 
& CheXpert         & Radiology      & 668   & \tick  \\ 
& US-Derm MCQA     & Dermatology    & 1,996     & -      \\ 
& Path MCQA        & Histopathology & 450      & -      \\ 
& EyePACS          & Ophthalmology  & 3,161     & -      \\ 
\hline
\multirow{2}{*}{Medical visual question-answering}
& MedXpertQA       & General medical & 2,000   & \tick \\ 
& SLAKE (English-only)   & Radiology   & 1,061    & -   \\
& VQA-RAD                & Radiology   & 2,248    & -   \\
\hline
\multirow{1}{*}{Chest X-ray report generation}
& MIMIC-CXR        & Radiology      & 306  & -  \\
\hline
\multirow{2}{*}{Medical agentic behavior}
& AgentClinic-MedQA    & Agentic text & 215  & -      \\
& AgentClinic-MIMIC-IV & Agentic text & 200  & \tick  \\
\hline
\multirow{3}{*}{General purpose}
& MMLU Pro             & Text         & 12,032  & N/A  \\
& Global MMLU Lite     & Text         & 6,400  & N/A  \\
& MMMU (val)           & Text+Image   & 900  & N/A  \\
\Xhline{2.5\arrayrulewidth}            
\end{tabular}
\begin{tablenotes}
  \scriptsize 
  \item $\dagger$ Out of Distribution: Data not seen during any model development stages. For general purpose benchmarks, it is difficult to determine if data are OOD given the large amount of pretraining data in the original mixture.
\end{tablenotes}

\end{table}

\subsection{General evaluation approach} \label{sssec:geneval}

\paragraph{Evaluation parameters:} Unless reported otherwise, all evaluations that we performed consisted of a single inference run per example. For MedGemma evaluations, a temperature of 0.0 was used on medical benchmarks and the default temperature was used on non-medical benchmarks. For evaluations of all other models, on all datasets, each model's default temperature and top-k were used. Due to data privacy and license terms, only publicly available datasets were used in evaluating models involving public APIs (e.g. OpenAI models). In cases where existing literature with performance metrics was available, those values were used and noted (with inclusion criteria for external models described below). For generalist models, we found that giving them a persona in the system message, such as  ``You are a helpful medical assistant'' or ``You are a helpful radiology assistant'' could improve their performance, we thus added these messages into the evaluation prompts. Detailed prompt usages can be found in Appendix Tables~\ref{tab:text_eval_prompts} and~\ref{tab:classification_eval_prompts}. As the DeepSeek R1 model~\cite{deepseekai2025deepseekr1incentivizingreasoningcapability} is a text-only model, it was only evaluated on text benchmarks.

\paragraph{Inclusion of previously published model performance data:} Where comparisons to other models were made, we restricted inclusion to models that met the following criteria: a publicly accessible model card (e.g., via Hugging Face or an institutional website); clear and verifiable attribution, including the name and contact information of the responsible individual or institution; explicit licensing terms governing its use; at least one associated technical report or publication for the current or a prior version of the model. These inclusion criteria were chosen to help ensure accountability, transparency, and adherence to sound machine learning practices, such as avoiding test dataset leakage. Additionally, for visual question answering (VQA) comparisons, we only include zero-shot generative evaluations for the most direct and meaningful comparisons (thus excluding few-shot or discriminative, embedding-based approaches for VQA).

\subsection{Medical text question-answering}

For evaluation of medical and health related capabilities, we used the official, publicly available test splits for MedQA, MedMCQA, PubMedQA, MMLU medical subcategories, AfriMed-QA, and MedXpertQA~\cite{medxpertqa}. AfriMed-QA includes a mix of closed and open questions though we limited evaluations to only the closed multiple choice questions (MCQs). No data from MedXpertQA was used in model training, so it is considered an \mbox{out-of-distribution} (OOD) benchmark.

\subsection{Medical image classification}
\label{sec:gemma-classification}

We evaluated medical image classification on three public chest X-ray datasets, as well as private datasets of dermatology, histopathology and retinal fundus images. We measured classification performance using accuracy or macro F1, depending on the dataset, targeting common findings or diagnoses in each modality.

\paragraph{Chest X-rays:} Prediction accuracy was evaluated for five conditions in the MIMIC-CXR \cite{johnson2019mimic, johnson2019mimicdatabase, goldberger2000physiobank} and CheXpert~\cite{irvin2019chexpert} datasets: atelectasis, cardiomegaly, consolidation, edema, and pleural effusion. Accuracy was evaluated for three conditions in the ChestX-ray14 (CXR14)~\cite{wang2017chestx} dataset: lung opacity, pneumothorax, and fracture. We used the original condition labels from CheXpert. For MIMIC-CXR, we report performance on two different versions of the test data: (1) using radiologist-adjudicated labels with missing and uncertain labels excluded, as in~\citet{yang2024advancing} and (2) using original labels from~\citet{MIMIC-CXR-JPG2019} with missing and uncertain labels considered to be negative, with the same set of test cases as reported in~\citet{hyland2023maira}. For the CXR14 data set, radiologist-adjudicated labels as described in~\citet{majkowska2020chest} were used for evaluation. Evaluations on MIMIC-CXR were not performed with the OpenAI o3 model due to data privacy considerations.

\paragraph{Dermatology:} ``US-Derm MCQA''~\cite{liu2020deep} is an internal, de-identified dataset consisting of one image per patient from 1996 patients who were referred to tele-dermatologists by primary care physicians in the United States. There are 136 different skin conditions represented across the images, with ground truth diagnoses provided by dermatologists based on the images and metadata. We converted this dataset into a multiple choice question format where the associated reference condition is included among three other randomly assigned condition labels from the same dataset (four options total per MCQ).

\paragraph{Histopathology:} ``Path MCQA'' is an internal dataset of 450 patches extracted from 354 unique whole slide images associated with specimens from breast cancer, lung cancer, prostate cancer, lymph nodes, and cervical biopsies. These represent test splits (by patient) from several different data sources. Patches comprise magnifications of 5x, 10x, and 20x (2, 1, and 0.5 microns per pixel). A single, multiclass labeling task for each patch was formulated as a multiple choice question, with four to nine possible options per question depending on the tissue type and labeling task associated with the image. These questions focus on identification and grading for breast cancer, prostate cancer, and cervical dysplasia as well as lung cancer histologic classification and histologic sub-typing. Ground truth labels were obtained via annotations provided as region-level labels by US board-certified pathologists~\cite{jaroensri2022deep,nagpal2019development,nagpal2020development, sadhwani2021comparative}. 

\paragraph{Ophthalmology:} We used the de-identified 45-degree fundus imagery dataset from EyePACS~\cite{cuadros2009eyepacs} as described previously~\cite{yang2024advancing}. We evaluated one image per patient from 3161 patients on clinically-determined 5-class diabetic retinopathy (DR) severity labels, with each task formulated as a multiple choice question with five options: none, mild, moderate, severe or proliferative DR.

\subsection{Medical visual question-answering}

We measured visual question-answering performance on the radiology SLAKE and VQA-RAD data sets using the average tokenized F1 metric across open and closed QAs, as well as accuracy on the subset of yes/no questions. For SLAKE, we used the default train/test split, and for VQA-RAD, we used splits from \citet{yang2024advancing}\footnote{\url{https://github.com/Google-Health/google-health/blob/master/data\_splits/}} to avoid the train/test image contamination present in the original splits.

\subsection{Chest X-ray report generation}

The MedGemma 4B pretrained model was used to generate radiology reports from the MIMIC-CXR test set. The pretrained model was chosen here rather than the post-trained model due to the sensitivity to reporting style of metrics like RadGraph F1~\cite{jain2021radgraph}. The pretrained model could better follow the style of MIMIC-CXR, as MIMIC-CXR reports are used in training, while the post-trained model conformed more closely to the original Gemma 3 style in terms of report generation.

We measured accuracy of chest X-ray report generation on the MIMIC-CXR dataset by comparing MedGemma-generated reports to the original radiologist reports for both impression and findings using the RadGraph F1 metric~\cite{jain2021radgraph} on the 912 image set used in~\citet{tanno2024consensus} and \citet{yang2024advancing}.

We also performed a human expert evaluation on the same 306-case image set as in~\citet{yang2024advancing} with a US board-certified cardiothoracic radiologist to evaluate both the original and MedGemma-generated reports with respect to the associated chest X-ray image. The evaluation task compared the reports on a five-point scale, as originally described in~\citet{yang2024advancing} and shown in Appendix Table~\ref{tab:human_report_rubric}. Although the reviewer was asked to remain neutral in their evaluation, they were not blinded to which report was from the original radiologist vs. from the AI system. This evaluation complements the automated RadGraph-based evaluation as it can both distinguish between major and minor issues as well as account for scenarios in which the original MIMIC-CXR report contains errors or omissions.

\subsection{Medical agentic behavior}

To provide insights into the capabilities of MedGemma in more complex environments, we measured the ability of MedGemma to operate in an agentic setting. We evaluated on AgentClinic \cite{schmidgall2024agentclinic}, which positions MedGemma in the role of a ``physician agent'' in a simulated clinical environment. Solving tasks in AgentClinic required the model to perform dialogue-driven patient history taking, ordering and interpreting medical exams, and operating under incomplete information in order to accurately provide a final diagnosis. We evaluated on 415 simulated cases using the text-only environments, AgentClinic-MedQA (215 cases) and AgentClinic-MIMIC-IV (200 cases), which are derived from MedQA \cite{medqa} and MIMIC-IV \cite{johnson2023mimic} respectively. 

\subsection{General purpose benchmarks}

Given the limitations that many specialized medical models exhibit when faced with non-medical tasks, we also evaluated possible tradeoffs of specialization by evaluating on the MMLU Pro, Global MMLU Lite, and MMMU benchmarks. Performance on the MMMU benchmark is reported on the validation set as the public test set does not include answers.

\section{MedGemma Results}
\label{sec:results_medgemma}
\paragraph{Medical text question-answering:} 
Across all text-only biomedical QA tasks evaluated, MedGemma demonstrated superior performance over the standard Gemma 3 model variant of the same size, as well as competitive performance with much larger models in many cases. This was based on evaluation on MedQA, MedMCQA, PubMedQA, MMLU subsets and AfriMed, shown in \cref{tab:results_text_transposed_revised}, and the out-of-distribution MedXpertQA, shown in \cref{tab:medXpert}. 

\begin{table}[htbp]
\centering
\caption{\small\textbf{Accuracy on text-only medical benchmarks.} Metrics for MedQA correspond to the original 4 option test set, unless otherwise specified with an asterisk. Metrics for external, small models as well as OpenBIOLLM 70B were obtained from the prior reports for these models. Other metrics were computed internally as described in the evaluation section.}
\label{tab:results_text_transposed_revised}
\resizebox{\textwidth}{!}{
\begin{tabular}{@{}lc|ccccccccccc@{}}
\toprule
Model \textsuperscript{$\dagger$} & \textbf{Open} \textsuperscript{$\ddagger$} & \makecell{MedQA} & MedMCQA & PubMedQA & \makecell{MMLU \\ \scriptsize{Anatomy}} & \makecell{MMLU \\ \scriptsize{Clinical Knowl.}} & \makecell{MMLU \\ \scriptsize{College Biology}} & \makecell{MMLU \\ \scriptsize{College Medicine}} & \makecell{MMLU \\ \scriptsize{Medical Genetics}} & \makecell{MMLU \\ \scriptsize{Prof. Medicine}} & \makecell{MMLU \\ \scriptsize{Virology}} & AfriMed \\
\midrule
\multicolumn{13}{@{}l}{\textbf{Small Models}} \\
\midrule
MedGemma 4B & \checkmark & 64.4 & 55.7 & 73.4 & 59.3 & 71.3 & 70.8 & 65.3 & 83.0 & 76.8 & 53.0 & 52.0 \\
Gemma 3 4B & \checkmark & 50.7 & 45.4 & 68.4 & 54.1 & 69.8 & 77.8 & 63.0 & 74.0 & 65.4 & 42.8 & 48.0 \\
\makecell[l]{MedGemma 27B \vspace{-2mm}\\ \scriptsize{(with test-time scaling)}} & \checkmark & \textbf{87.7} & \textbf{74.2} & \textbf{76.8} & \textbf{83.7} & \textbf{86.0} & \textbf{96.5} & \textbf{86.1} & \textbf{97.0} & \textbf{93.4} & \textbf{53.6} & \textbf{84.0} \\
Gemma 3 27B & \checkmark & 74.9 & 62.6 & 73.4 & 74.8 & \textbf{86.0} & 93.8 & 78.6 & 91.0 & 85.7 & 51.2 & 72.0 \\
\makecell[l]{BioMistral DARE 7B \textsuperscript{\S} \vspace{-2mm}\\ \scriptsize~\cite{labrak2024biomistral}} & \checkmark & 
51.1 & 48.7 & 77.7 & 55.8 & 62.3 & 66.9 & 58.0 & 67.0 & 61.4 & N/A & N/A \\
\makecell[l]{JSL-MedLlama 3 8B v2.0 \textsuperscript{\S} \vspace{-2mm}\\ \scriptsize~\cite{jsl_medllama_v2_8b}} & \hphantom{\textsuperscript{||}}\checkmark\textsuperscript{||} & 61.4\textsuperscript{*} & 61.2 & 74.2 & 71.9 & 78.1 & 82.6 & 71.1 & 83.0 & 78.7 & N/A & N/A \\
\makecell[l]{OpenBioLLM 8B \textsuperscript{\S} \vspace{-2mm}\\ \scriptsize~\cite{OpenBioLLMs}} & \checkmark & 59.0 & 56.9 & 74.1 & 69.8 & 76.1 & 84.2 & 68.0 & 86.1 & 78.2 & N/A & N/A \\
\makecell[l]{IQVIA Med-R1 8B \textsuperscript{\S} \vspace{-2mm}\\ \scriptsize~\cite{iqvia_2025_med-r1}} & - & 73.3 & 63.3 & 76.4 & 72.6 & 78.5 & 88.2 & 72.8 & 87.0 & 84.9 & N/A & N/A \\
\midrule
\multicolumn{13}{@{}l}{\textbf{Large Models}} \\
\midrule
\makecell[l]{OpenBioLLM 70B \textsuperscript{\S} \vspace{-2mm}\\ \scriptsize~\cite{OpenBioLLMs}} & \checkmark & 78.2 & 74.0 & 79.0 & 83.9 & 92.9 & 93.8 & 85.7 & 93.2 & 93.8 & N/A & N/A \\
\makecell[l]{DeepSeek R1 \vspace{-2mm}\\ \scriptsize~\cite{deepseekai2025deepseekr1incentivizingreasoningcapability}} & \checkmark & 90.1 & 78.8 & 77.2 & 91.1 & 91.7 & \textbf{98.6} & \textbf{90.8} & 99.0 & 95.6 & 56.0 & \textbf{92.0} \\
Gemini 2.5 Flash & - & 92.0 & 79.7 & 76.2 & 91.1 & 91.7 & \textbf{98.6} & 87.9 & 97.0 & 96.0 & \textbf{59.6} & 84.0 \\
Gemini 2.5 Pro   & - & 92.6 & 81.1 & 75.8 & 91.1 & 91.7 & \textbf{98.6} & 89.0 & 96.0 & \textbf{96.3} & 56.0 & 84.0 \\
GPT-4o & - & 86.5 & 76.1 & 78.4 & 86.7 & 89.4 & 94.4 & 86.7 & 98.0 & 93.0 & 58.4 & 80.0 \\
o3     & - & \textbf{93.3} & \textbf{83.3} & \textbf{80.0} & \textbf{91.9} & \textbf{94.7} & \textbf{98.6} & 90.2 & \textbf{100}  & 96.0 & 56.0 & 84.0 \\
\bottomrule
\end{tabular}
}
\begin{tablenotes}
  \scriptsize 
  \item * Indicates results for the all-options version of the dataset
  \item $\dagger$ See Section~\ref{sssec:geneval} for inclusion criteria
  \item $\ddagger$ Open-weight models
  \item $\S$ Prior reported results
  \item || Research use only
\end{tablenotes}
\end{table}

\paragraph{Medical image classification:} To further evaluate MedGemma across the modalities that are most highly represented in the training data, we utilized a set of image-based classification tasks across radiology, histopathology, dermatology, and ophthalmology. Although addressing classification as a zero-shot generative task may not provide maximum performance compared to training an embedding-based classifier, these evaluations provide additional insights into the quality and performance of underlying image encoder and the overall model capabilities. Results are summarized in \cref{tab:results_zeroshot_updated} for CXR evaluations and \cref{tab:results_zeroshot_others} for histopathology, dermatology, and retina image classification. MedGemma demonstrated substantially superior performance on these tasks relative to the Gemma 3 baselines, and notably, MedGemma also demonstrated superior performance as compared to much larger API-based models.

\begin{table}[htbp]
\centering
\caption{\small{\textbf{Accuracy results on MedXpertQA (OOD)}}}
\label{tab:medXpert}
\resizebox{\textwidth}{!}{
\begin{tabular}{l|c|c|c|c|c|c|c}
\Xhline{2.5\arrayrulewidth}
\multicolumn{1}{l|}{} & \multicolumn{4}{c|}{\textbf{Small Models}} & \multicolumn{3}{c}{\textbf{Large Models}} \\
\cmidrule(lr){1-8}
Type & \makecell{MedGemma 4B} & \makecell{Gemma 3 4B}  & \makecell{MedGemma 27B} & \makecell{Gemma 3 27B} & \makecell{Gemini 2.5 Flash} & \makecell{Gemini 2.5 Pro} & \makecell{o3}\\
\Xhline{2.5\arrayrulewidth}
\multirow{1}{*}{Text-only} & 14.2 & 11.6 & \textbf{25.7} & 15.7 & 36.2 & 43.1 & \textbf{54.6} \\
\hline
Multi-modal only & 24.4 & 22.3 & N/A & \textbf{29.8} & 47.4 & 58.9 & \textbf{67.5} \\
\Xhline{2.5\arrayrulewidth}
\end{tabular}
}
\end{table}

MedGemma is also capable of engaging in open-ended clinical reasoning tasks. Examples of such tasks with MedGemma 27B, including clinical commentary on model performance, are shown in Table~\ref{tab:oncology_text_example_long} (long-form MedGemma response) and Table~\ref{tab:oncology_text_example_succinct} (succinct MedGemma response). Additional examples can be found in Appendix Section~\ref{sec:additional_med_reasoning_examples}.

\begin{table}[htbp]
\caption{Example of MedGemma 27B clinical reasoning in an oncology setting, including commentary by a US board-certified practicing oncologist}
\label{tab:oncology_text_example_long}
\renewcommand{\arraystretch}{1.1}
\setlength{\tabcolsep}{4pt}
\footnotesize

\begin{tabularx}{\textwidth}{@{} >{\raggedright\arraybackslash}X @{}}
\toprule
\textbf{Prompt} \\
\midrule
A 62-year old man presents with a 2 cm spiculated right upper lobe peripheral lung nodule and mediastinal lymphadenopathy. Imaging furthermore shows multiple metastatic lesions in the liver and bones, and a solitary 1 cm lesion in the right parietal lobe without surrounding vasoedema. A biopsy from the liver demonstrated adenocarcinoma consistent with a lung primary. Genomic sequencing demonstrated TP53 and EGFR L858R mutation, and PD-L1 was 70\%. On examination, the patient is without any neurologic deficits and is ECOG 0. What is the best next step in management for this patient? \\
\midrule
\textbf{MedGemma 27B output} \\
\midrule
Okay, let's break down the management for this patient with Stage IV NSCLC.

\textbf{Key Findings:}
\begin{enumerate}[nosep, leftmargin=*]
    \item \textbf{Diagnosis:} Stage IV Adenocarcinoma of the Lung (confirmed by liver biopsy).
    \item \textbf{Driver Mutation:} EGFR L858R mutation (highly actionable).
    \item \textbf{Biomarker:} PD-L1 70\% (high expression).
    \item \textbf{Metastases:} Widespread (mediastinal nodes, liver, bone, solitary brain lesion).
    \item \textbf{Clinical Status:} Excellent performance status (ECOG 0), asymptomatic brain lesion.
\end{enumerate}

\textbf{Analysis \& Rationale:}
\begin{itemize}[nosep, leftmargin=*]
    \item \textbf{EGFR Mutation is Paramount:} The presence of the EGFR L858R mutation is the most critical factor guiding first-line therapy. Patients with EGFR-mutated NSCLC have demonstrated significantly improved outcomes with EGFR tyrosine kinase inhibitors (TKIs) compared to chemotherapy.
    \item \textbf{PD-L1 High:} While PD-L1 is high (70\%), indicating potential benefit from immunotherapy, current guidelines and clinical trial data (like FLAURA) strongly favor EGFR TKIs as the \textit{first-line} treatment for patients with EGFR mutations, even those with high PD-L1. Immunotherapy is typically considered after TKI failure or in combination with chemotherapy in specific settings, but not usually as the initial monotherapy in this scenario.
    \item \textbf{Brain Metastasis:} The solitary, asymptomatic brain lesion needs management. EGFR TKIs, particularly third-generation agents like Osimertinib, have good central nervous system (CNS) penetration and activity. Therefore, systemic therapy with an EGFR TKI is often the initial approach to treat the brain metastasis as well. Close monitoring with MRI is essential. If the lesion progresses despite systemic therapy or becomes symptomatic, Stereotactic Radiosurgery (SRS) would be considered.
    \item \textbf{ECOG 0:} The patient's excellent performance status makes him a suitable candidate for aggressive systemic therapy.
\end{itemize}

\textbf{Best Next Step in Management:}
The best next step is to initiate \textbf{first-line systemic therapy with an EGFR tyrosine kinase inhibitor (TKI)}.

\begin{itemize}[nosep, leftmargin=*]
    \item \textbf{Preferred Agent:} \textbf{Osimertinib} is generally the preferred first-line EGFR TKI due to its superior efficacy (progression-free survival and overall survival) and improved CNS penetration compared to earlier-generation TKIs (like Gefitinib, Erlotinib, Afatinib) demonstrated in the FLAURA trial.
    \item \textbf{Management of Brain Metastasis:} Start Osimertinib. Monitor the brain lesion closely with serial MRI scans (e.g., every 3-6 months initially). Consider SRS if the lesion progresses or causes symptoms.
    \item \textbf{Supportive Care:} Address potential side effects of the TKI (e.g., rash, diarrhea). Manage potential bone pain with analgesics or palliative radiation if needed.
\end{itemize} \\
\midrule
\textbf{Commentary by US board-certified practicing oncologist} \\
\midrule
\textbf{Strengths:} Identifies the key components of staging and the actionable driver mutation. PD-L1 is high but correctly identifies TKI as preferable over immunotherapy. Correctly identifies osimertinib as having good brain penetration especially with asymptomatic and small brain lesions. Appropriately highlights FLAURA trial. Includes supportive care aspects including potential side effects.

\textbf{Areas for Improvement:} Immunotherapy can be considered after TKI progression, though the treatment paradigm has evolved (i.e., amivantamab + chemotherapy, datopotamab deruxtecan). Although osimertinib is a correct answer, newer combinations are available and have shown benefit over osimertinib alone (i.e., amivantamab + lazertinib, osimertinib + chemotherapy). Serial brain MRI is mentioned but serial rest-of-body imaging is not. Dosing of osimertinib not mentioned. Bone disease discussed, though no consideration for bone modifying agents or for whether surgery may be indicated. Shared decision-making would be a beneficial addition.\\
\bottomrule
\end{tabularx}
\end{table}


\begin{table}[htbp]
\caption{Example of MedGemma 27B clinical reasoning in an oncology setting, specifically prompted for \emph{succinct} output, including commentary by a US board-certified practicing oncologist}
\label{tab:oncology_text_example_succinct}
\renewcommand{\arraystretch}{1.1}
\small

\begin{tabularx}{\textwidth}{@{} >{\raggedright\arraybackslash}X @{}}
\toprule
\textbf{Prompt} \\
\midrule
A 62-year old man presents with a 2 cm spiculated right upper lobe peripheral lung nodule and mediastinal lymphadenopathy. Imaging furthermore shows multiple metastatic lesions in the liver and bones, and a solitary 1 cm lesion in the right parietal lobe without surrounding vasoedema. A biopsy from the liver demonstrated adenocarcinoma consistent with a lung primary. Genomic sequencing demonstrated TP53 and EGFR L858R mutation, and PD-L1 was 70\%. On examination, the patient is without any neurologic deficits and is ECOG 0. What is the best next step in management for this patient? \emph{Please be succinct.} \\
\midrule
\textbf{MedGemma 27B output} \\
\midrule
Initiate Osimertinib and monitor the brain lesion with serial MRI. \\
\midrule
\textbf{Commentary by US board-certified practicing oncologist} \\
\midrule
\textbf{Strengths:} To the point, directly addresses systemic therapy + serial brain MRI.

\textbf{Areas for Improvement:} A little too succinct, does not discuss radiosurgery, and otherwise the same issues as the lengthy output in the prior example. \\
\bottomrule
\end{tabularx}
\end{table}

\begin{table}[htb]
\centering
\caption{\small\textbf{Medical image classification with zero-shot generative output for chest X-ray.} This table presents the performance on chest X-ray classification for presence of specific conditions, evaluated as a 0-shot, generative task. For MIMIC-CXR and CheXpert datasets, macro F1 is calculated for atelectasis, cardiomegaly, consolidation, edema, and pleural effusion. For CXR14, macro F1 is calculated for lung opacity, pneumothorax, and fracture. }
\label{tab:results_zeroshot_updated}
\resizebox{\textwidth}{!}{
\begin{tabular}{@{}ll|ccc|ccc|c@{}}
\toprule
\multicolumn{2}{c|}{} & \multicolumn{3}{c|}{\textbf{Small Models}} & \multicolumn{3}{c|}{\textbf{Large Models}} & \\
\cmidrule(lr){3-8}

\multirow{1}{*}{\makecell[tl]{Dataset}} &
\multirow{1}{*}{\makecell[tl]{Metric}} &
\textbf{MedGemma 4B} & \textbf{Gemma 3 4B} & \textbf{Gemma 3 27B} & \makecell[c]{\textbf{Gemini v2.5} \\ \textbf{Flash}} & \makecell[c]{\textbf{Gemini v2.5} \\ \textbf{Pro}} &
\textbf{o3} & SOTA VLM \textsuperscript{$\S$} \\

\midrule

\makecell[l]{MIMIC-CXR \\ \small{Med-Gemini test set\textsuperscript{*}}} & \multirow{2}{*}{\makecell[c]{macro F1 \\ (5 conditions)}} & \textbf{88.9} & 81.2  & 71.7 & 81.0 & \textbf{85.8} & N/A & \makecell[c]{\textbf{90.7} \\ Med-Gemini \\ \cite{yang2024advancing}} \\
\makecell[l]{MIMIC-CXR \\ \small{MAIRA test set}} & & \textbf{40.5} & 26.7 & 25.0 & \textbf{32.2} & 31.9 & N/A & \makecell[c]{\textbf{51.6} \\ Med-PaLM M\\\cite{tu2024towards}}\\
\midrule
CheXpert (OOD) & \makecell[c]{macro F1 \\ (5 conditions)} & \textbf{48.1} & 32.6 & 26.2 & 37.4 & 37.0 & \textbf{40.9} & \makecell[c]{\textbf{49.0}\textsuperscript{\textdagger} \\ RadVLM \\ \cite{deperrois2025radvlm}} \\
CXR14 (OOD)\textsuperscript{*} & \makecell[c]{macro F1 \\ (3 conditions)}  & \textbf{50.1} & 32.0 & 31.4 & 36.6 & \textbf{39.2} & 32.0 & \makecell[c]{46.7 \\ Med-Gemini \\ \cite{yang2024advancing}}\ \\

\bottomrule
\end{tabular}
}
\begin{tablenotes}
  \scriptsize

\item * Radiologist adjudicated labels are used \cite{yang2024advancing} Section A.1.1., so metrics may not be directly comparable to those reported in the literature for external models.
\item \textdagger \ RadVLM macro F1 was calculated based on the individual F1 scores reported, estimated from the associated bar chart
\item $\S$ Prior reported results  
\end{tablenotes}

\end{table}

\begin{table}[htb]
\centering
\caption{\small\textbf{Medical image classification with zero-shot generative output for histopathology, dermatology, and retina.} Performance on internal classification tasks for histopathology, dermatology, and retinal fundus images; formulated as zero-shot, multiple choice tasks for a generative model.}
\label{tab:results_zeroshot_others}
\resizebox{\textwidth}{!}{
\begin{tabular}{@{}lll|ccc|cc@{}}
\toprule

\multicolumn{3}{c|}{} & \multicolumn{3}{c|}{\textbf{Small Models}} & \multicolumn{2}{c}{\textbf{Large Models}} \\
\cmidrule(lr){4-8}

\makecell[tl]{Modality} &
\makecell[tl]{Dataset} &
\makecell[tl]{Metric} &
\textbf{MedGemma 4B} & \textbf{Gemma 3 4B} & \textbf{Gemma 3 27B} & \makecell[c]{\textbf{Gemini v2.5} \\ \textbf{Flash}} & \makecell[c]{\textbf{Gemini v2.5} \\ \textbf{Pro}}\\
\midrule

Histopath. & PathMCQA & accuracy & \textbf{69.8} & 37.1 & 42.2 & \textbf{46.9} & 42.7 \\
\midrule

Skin & US-Derm MCQA & accuracy & \textbf{71.8} & 52.5 & 66.9 & 78.4 & \textbf{81.0} \\
\midrule

\multirow{1}{*}{\makecell[l]{Retina}} & \makecell[l]{EyePACS} & accuracy & \textbf{64.9} & 14.4 & 20.3 & 17.5 & \textbf{27.7} \\
\bottomrule
\end{tabular}
}
\end{table}

\begin{figure}[htp]
    \centering
    \includegraphics[width=1.0\linewidth]{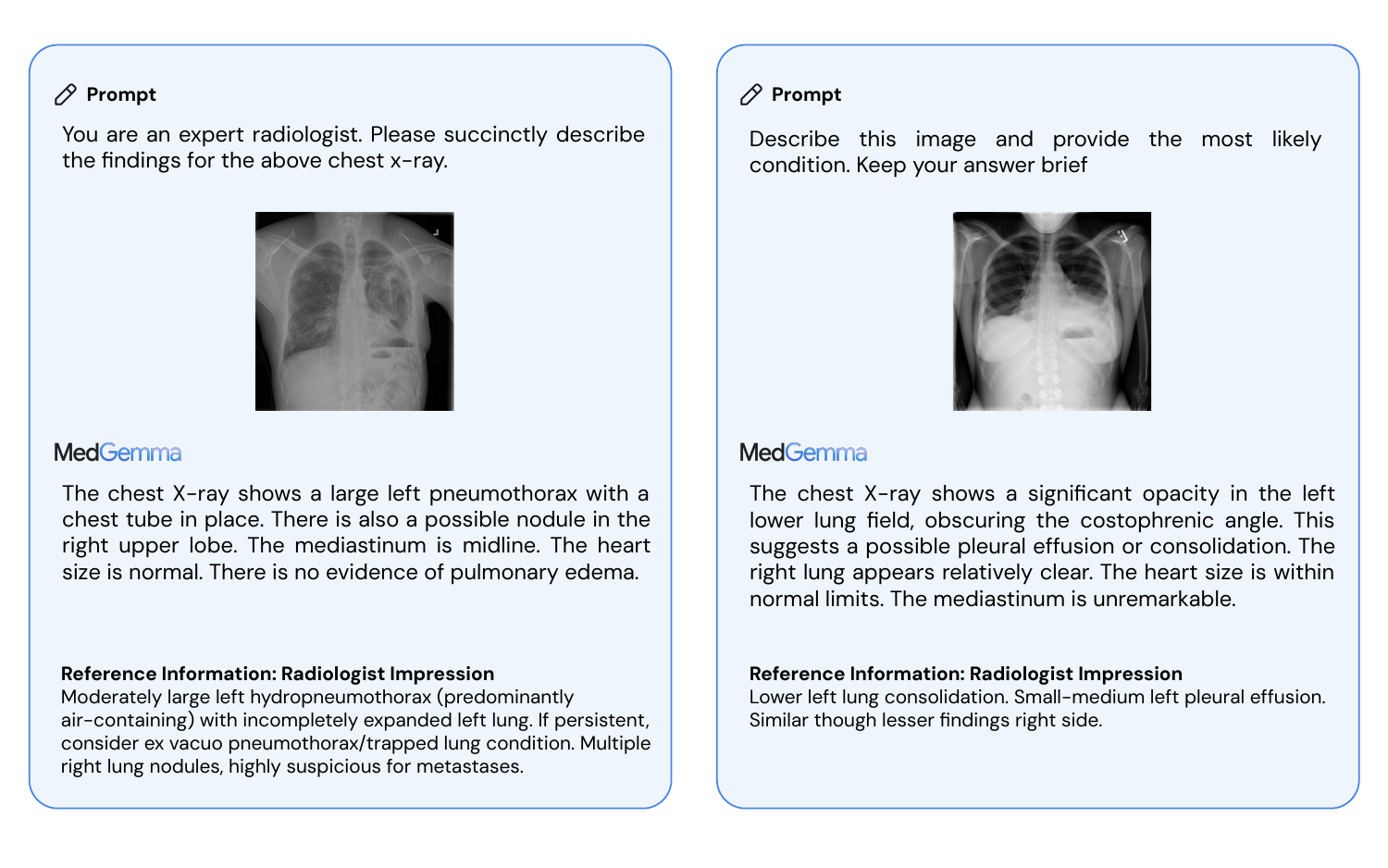}
    \includegraphics[width=1.0\linewidth]{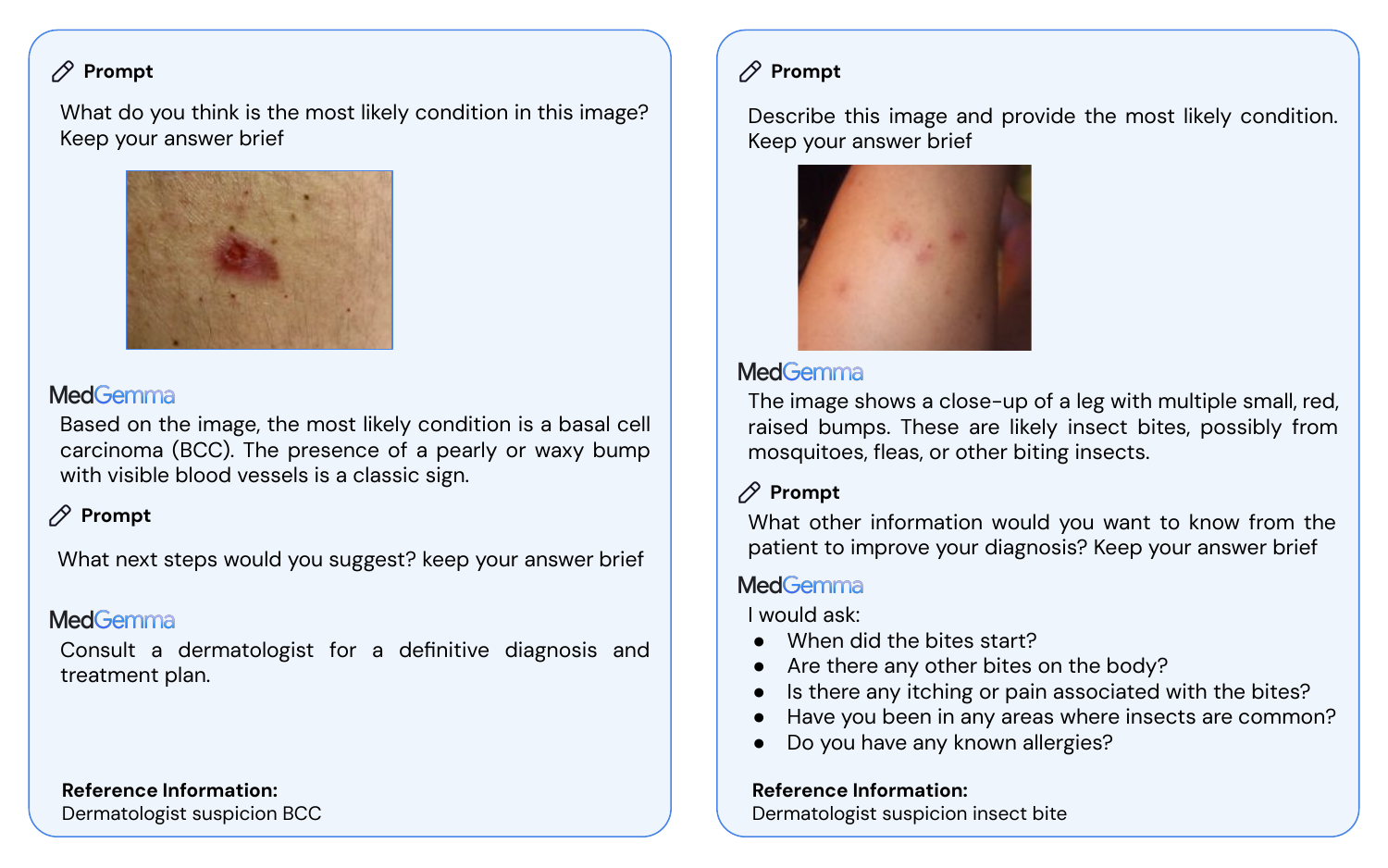}
    \caption{\small{\textbf{Example of CXR and dermatology image dialogue via open-ended question answering with MedGemma 4B.}}}
    \label{fig:anecdotal_cxr_derm}
\end{figure}

\begin{figure}[htp]
    \centering
    \includegraphics[width=1.0\linewidth]{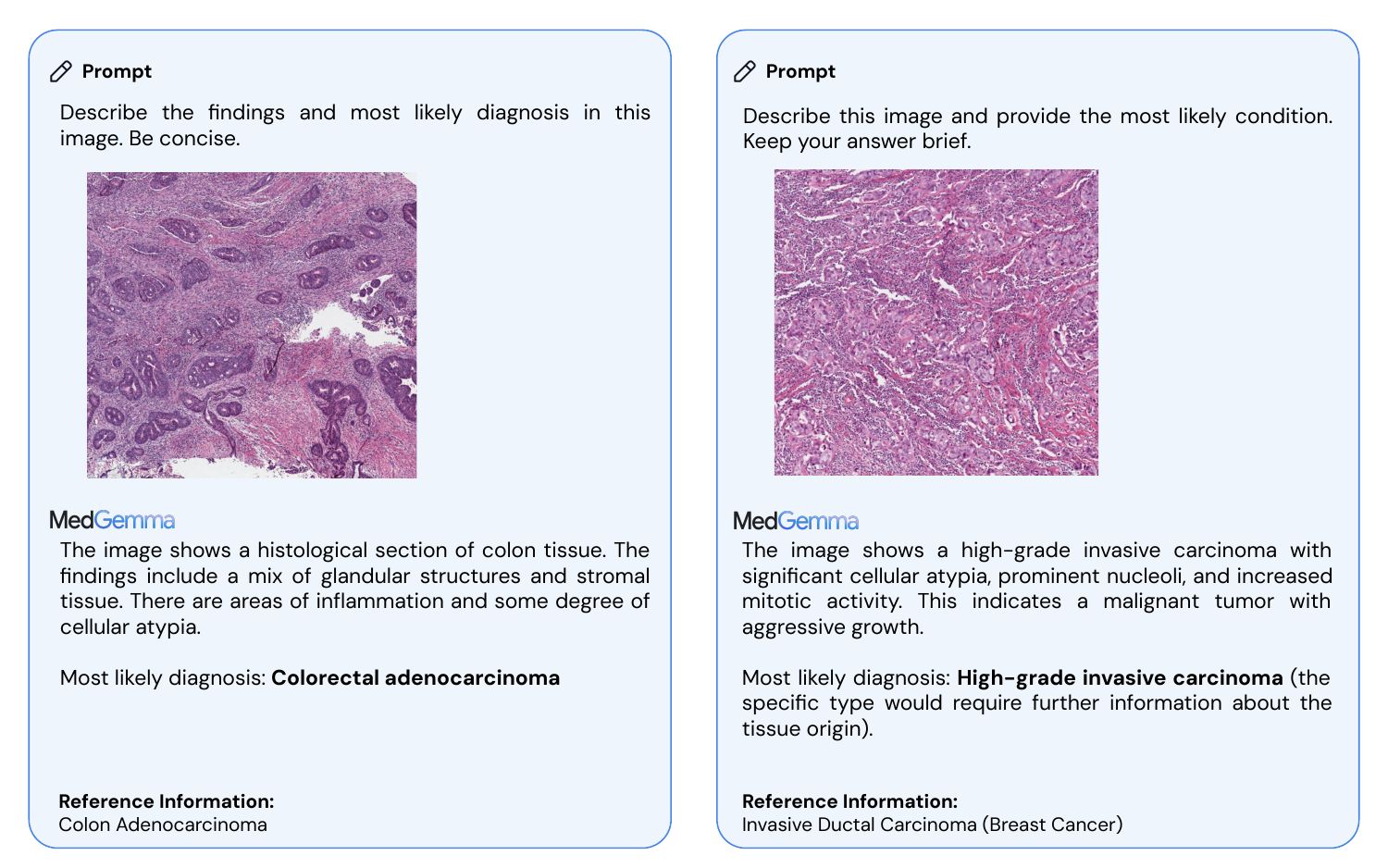}
    \caption{\small{\textbf{Example of histopathology open-ended question answering with MedGemma 4B.}}}
    \label{fig:anecdotal_path}
\end{figure}

\paragraph{Medical visual question answering:} 
Performance for SLAKE \cite{liu2021slake} and VQA-RAD \cite{lau2018dataset} are summarized in \cref{tab:results_mvqa_transposed_updated}. Again, MedGemma 4B demonstrated superior performance over the standard Gemma 3 model variant of the same size, as well as competitive performance with much larger models in many cases. Of note, we did not compare to models for which these metrics are calculated using a discriminative, embedding-based approach. Such approaches can be informative, but use the train set to define the set of possible text answers, and thus represent a classification task instead of a more flexible open-ended VQA task. Additionally,  LLaVA-Med and BiomedGPT-B are included in the results table using previously reported metrics \cite{li2023llava, zhang2023biomedgpt} and demonstrated very strong performance, with the qualifier that such performance involves supervised fine-tuning for the specific datasets. Additionally, LLaVA-Tri~\cite{xie2024medtrinity} has reported even higher performance metrics, but the reported methodological details for these metrics are insufficient to inform direct comparison.
In addition to these statistical results, we show several anecdotal examples of open-ended visual question-answering in Figures~\ref{fig:anecdotal_cxr_derm} and~\ref{fig:anecdotal_path}.

\begin{figure}[htp]
    \centering
    \includegraphics[width=0.8\linewidth]{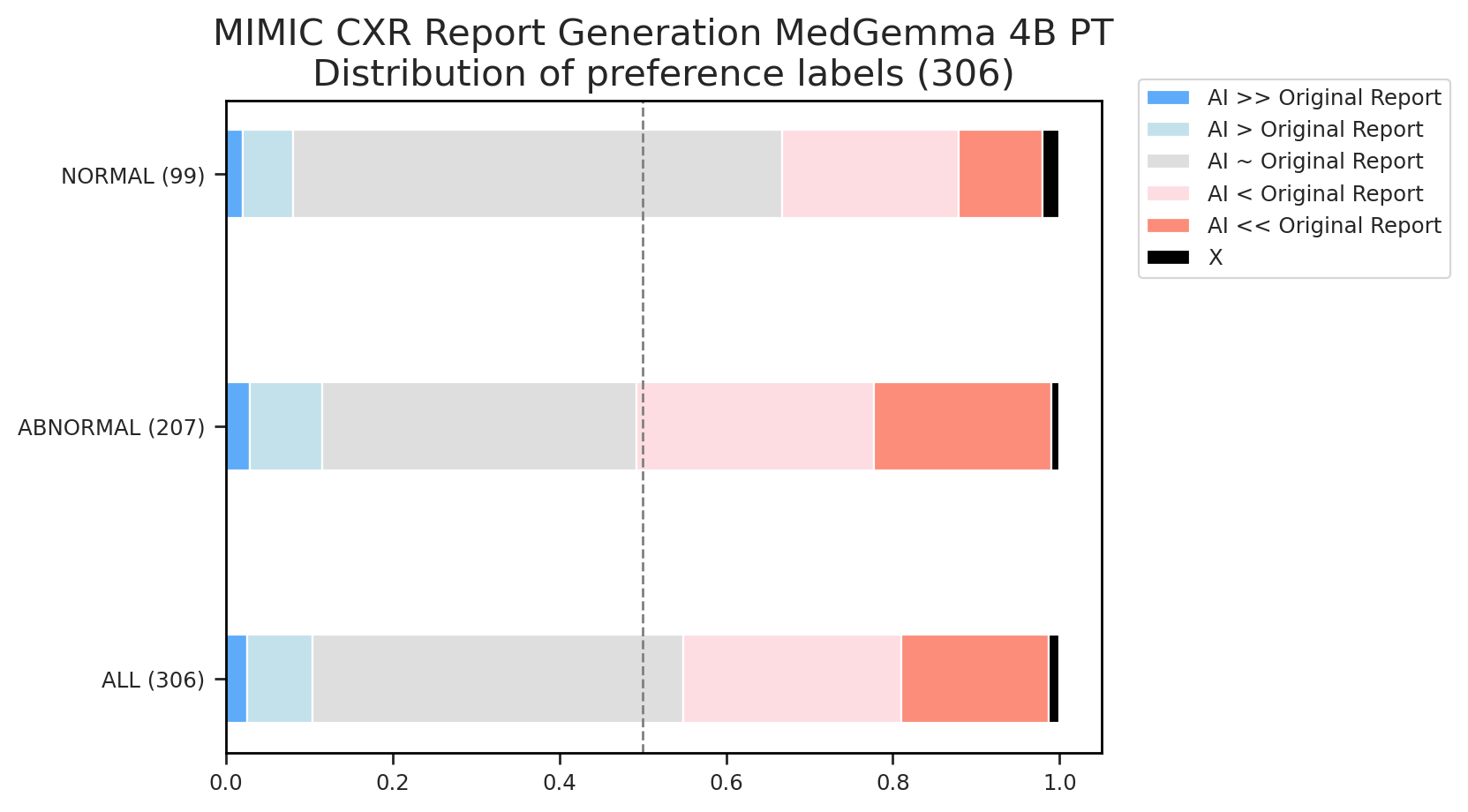}
    \caption{\small{The MedGemma 4B PT model was used to generate radiology reports on the MIMIC-CXR test set. A single board-certified thoracic radiologist reviewed the MIMIC-CXR report, generated report, and the corresponding CXR image to judge the quality of the reports. Images were reviewed using the original DICOMs on a clinical diagnostic viewer. Across all cases, 81\% of MedGemma’s CXR reports resulted in the same or superior clinical decision in comparison to the original reports. 
    }}
    \label{fig:medgemma_cxr_human}
\end{figure}

\paragraph{Chest X-ray report generation:}
Results of performance on CXR report generation based on the MIMIC-CXR dataset and the RadGraph F1 metric are shown in Table~\ref{tab:mimic-report-all-metrics}. The results demonstrate that our pretrained MedGemma models can achieve SOTA-level results by training with MIMIC-CXR dataset as our only CXR-related dataset. This shows the versatility of Gemma's architecture in adapting to complex medical use cases. In addition, expert human evaluation shown in Figure~\ref{fig:medgemma_cxr_human} found that 68\% and 49\% of generated reports in normal and abnormal studies respectively were equal or superior to the original reports, with 81\% of generated reports overall resulting in the same or superior clinical decisions in comparison to the original reports. For comparison, in a similar study, 73\% of reports generated by the larger Med-Gemini model would have resulted in the same or superior clinical decisions in comparison to the original reports \cite{yang2024advancing}.

\paragraph{Medical agentic behavior:} To provide initial insights into the capabilities of MedGemma in the setting of more realistic use cases beyond single turn question answering, we also evaluated MedGemma in an agentic role using the AgentClinic benchmark \cite{schmidgall2024agentclinic}. In this task, the model functions as a ``physician agent'' in simulated clinical encounters, requiring tasks like patient history taking, exam interpretation, and diagnosis under uncertainty to reach correct answers. Using the text-only AgentClinic-MedQA and AgentClinic-MIMIC-IV environments (with GPT-4o for the patient agent and measurement), MedGemma 27B showed superior performance to Gemma 3 27B, exceeding performance of human physicians on AgentClinic-MedQA, and approaching performance of much larger models on both datasets. Results are summarized in Table \ref{tab:agentclinic_results}. Of note, the 4B variants (MedGemma and Gemma 3) were not well suited for this task, demonstrating difficulty following system instructions for the agentic framework as provided by AgentClinic.


\begin{table}[htbp]
\centering
\caption{\small\textbf{Medical VQA results for SLAKE and VQA-RAD.} We include comparisons to other generative models. Additionally, the original VQA-RAD test set includes some duplicated images in the train and test sets (with different questions). As such we have previously described our own splits to avoid this contamination, but precludes direct comparison to externally reported metrics for this dataset.}
\label{tab:results_mvqa_transposed_updated}
\resizebox{\textwidth*4/5}{!}{
\begin{tabular}{@{}l|ccc|cc@{}}
\toprule
& \multicolumn{3}{c|}{\textbf{SLAKE (English)}} & \multicolumn{2}{c}{\makecell{\textbf{VQA-RAD}  \\ Test split from~\citep{xu2023elixr}}} \\
\cmidrule(lr){2-4} \cmidrule(lr){5-6}
\textbf{Model} & \makecell{overall \\ token f1} & \makecell{open-ended \\ token recall} & \makecell{closed-ended \\ accuracy} & \makecell{overall \\ token f1} & \makecell{closed Q\&A \\ accuracy} \\ 
\midrule
\multicolumn{6}{@{}l}{\textbf{Small Models}} \\
\midrule
MedGemma 4B & 72.3 & 63.3 & 87.6 & \textbf{49.9} & \textbf{69.1} \\
Gemma 3 4B   & 40.2 & 33.3 & 53.0 & 33.6 & 48.7 \\
Gemma 3 27B  & 42.5 & 30.8 & 64.5 & 42.7 & 59.4  \\
BiomedGPT-B~\cite{zhang2023biomedgpt}\textsuperscript{$\S$}  & \textbf{85.2} & - & \textbf{89.9} & * & * \\
LLaVA-Med \small{(BioMedCLIP)}\cite{li2023llava}\textsuperscript{$\S$}  & - & \textbf{87.1} & 86.8 & * & * \\
\midrule
\multicolumn{6}{@{}l}{\textbf{Large Models}} \\
\midrule
Med-Gemini \cite{yang2024advancing} & \textbf{75.8} & \textbf{72.2} & \textbf{84.6} & 50.1 & 69.7  \\
Gemini 2.5 Flash  & 54.6 & 42.2 & 80.9 & 53.6 & 70.2 \\
Gemini 2.5 Pro  & 53.1 & 40.4 & 78.6 & \textbf{54.2} & 71.4 \\
o3 & 55.5 & 45.0 & 76.3 & 52.5 & \textbf{71.9} \\
\bottomrule
\end{tabular}
}
\begin{tablenotes}
  \footnotesize

  \item * Reported results not available for the same test split (only for the original VQA-RAD splits).
  \item - Results for this metric not available in associated report.
  \item $\S$ Prior reported results
\end{tablenotes}
\end{table}

\begin{table}[t!]
\centering
\small
\caption{\small\textbf{Automated report generation metrics on the MIMIC-CXR dataset.} This table presents the performance of various models on generating radiology reports for chest X-rays using the publicly available MIMIC-CXR dataset. The \emph{Sections} column indicates whether the model generates the FINDINGS (F) or IMPRESSION (I) section of the report, with metrics sourced from published research. For all of the metrics, higher is better. Bold values highlight the best results in each section. MedGemma addresses the more challenging task of creating both sections (F + I) for frontal chest X-rays (anterior-posterior or posterior-anterior views), aiming to capture the radiologist's comprehensive interpretation of the study. Note that reported metrics across all models may not be directly comparable due to differences in exclusion and inclusion criteria for the test set in the respective citations.}
\label{tab:mimic-report-all-metrics}
\begin{threeparttable}
\resizebox{\textwidth*2/3}{!}{
\renewcommand{\arraystretch}{1.15}
\begin{tabular}{l|c|c}
\toprule[2.5\arrayrulewidth] 
\multirow{2}{*}{\textbf{Model}} & \multirow{2}{*}{\textbf{Sections}} & \multicolumn{1}{c}{\textbf{ Clinical Metric (\%)}} \\ [0.2em] \cline{3-3}
& & RadGraph F1 \textsuperscript{\S} \\ [0.2em]\midrule[2\arrayrulewidth]
CXR-RePaiR \citep{endo2021retrieval}                    & F only & 9.1 \\
$\mathcal{M}^2$ Transformer \citep{miura2020improving}  & F only & 22.0 \\
Med-PaLM M, 12B \citep{tu2024towards}                   & F only & 25.2 \\
Med-PaLM M, 84B \citep{tu2024towards}                   & F only & 26.7 \\
MAIRA-1~\cite{hyland2023maira}                          & F only & 24.3 \\
MAIRA-2~\cite{bannur2024maira}                          & F only & \textbf{34.6} \\\midrule[2\arrayrulewidth]
R2Gen \citep{chen2020generating}                        & F + I & 13.4 \\
WCT \citep{yan2021weakly}                               & F + I & 14.3 \\
CvT-21DistillGPT2 \citep{nicolson2023improving}         & F + I & 15.4 \\
Flamingo-CXR \citep{tanno2024consensus}                 & F + I & 20.5 \\
Med-Gemini-2D \citep{yang2024advancing}                 & F + I & 24.4 \\
PaliGemma 2 10B~\cite{steiner2024paligemma}             & F + I & 29.5 \\
MedVersa \cite{zhou2024generalistlearnermultifacetedmedical}        & F + I & \textbf{30.0} \\
\midrule[2.5\arrayrulewidth]
MedGemma 4B PT                                             & F + I & 29.5 \\
\bottomrule[2.5\arrayrulewidth] 
\end{tabular}
} 
\begin{tablenotes} 
  \footnotesize
  \item $\S$ Prior reported results except for MedGemma
\end{tablenotes}
\end{threeparttable}
\end{table}

\begin{table}[htbp]
\centering
\caption{\small\textbf{Performance on AgentClinic benchmarks.}}
\label{tab:agentclinic_results}

\footnotesize 

\begin{tabular}{@{}lcc@{}}
\toprule
\textbf{Model} & \makecell{\textbf{AgentClinic-MedQA} \\ Accuracy(\%)} & \makecell{\textbf{AgentClinic-MIMIC (OOD)} \\ Accuracy(\%)}  \\
\midrule
\multicolumn{3}{@{}l}{\textbf{Small Models}} \\
\midrule
MedGemma 27B            & \textbf{56.2} & \textbf{46.0} \\
Gemma 3 27B              & 50.7 & 35.2 \\
\midrule
\multicolumn{3}{@{}l}{\textbf{Large Models}} \\
\midrule
DeepSeek R1\textbf{~\cite{deepseekai2025deepseekr1incentivizingreasoningcapability}}              & 58.1 & 43.8 \\
Gemini 2.5 Pro         & 58.3 & 48.4 \\
o3                     & \textbf{65.8} & \textbf{50.6} \\
\midrule
Human physician        & 54.0\textsuperscript{*} & N/A \\
\bottomrule
\end{tabular}%
\begin{tablenotes}
  \scriptsize

\item * Human metric based on GPT-4 patient agent as reported in~\citealt{schmidgall2024agentclinic}. All other metrics were recalculated using GPT-4o as patient agent to minimize deviation from original AgentClinic report. 

\end{tablenotes}
\end{table}

\paragraph{General purpose benchmarks:}

Given the limitations and poor performance that many medically-specialized models exhibit when faced with non-medical tasks, we also evaluated for possible tradeoffs of specialization. Models were evaluated on MMLU Pro, Global MMLU Lite, and MMMU benchmark datasets, with performance comparisons between MedGemma and Gemma 3 variants shown in (\cref{tab:model_comparison}). MedGemma demonstrated only minor decreases in performance relative to the general models of the same size, further suggesting potential utility for downstream applications that require both specialized as well as generalized capabilities such as instruction following or diverse user interactions.

\begin{table}[htbp]
\centering
\caption{\small{\textbf{Accuracy results on general, non-medical benchmarks.}}}
\label{tab:model_comparison}
\resizebox{\textwidth}{!}{
\begin{tabular}{l|l|c|c|c|c}
\Xhline{2.5\arrayrulewidth}
Type & Benchmark & \makecell{MedGemma 4B} & \makecell{Gemma 3 4B\textsuperscript{\S}}  & \makecell{MedGemma 27B} & \makecell{Gemma 3 27B\textsuperscript{\S}}\\
\Xhline{2.5\arrayrulewidth}
\multirow{2}{*}{Text-only} & MMLU Pro & 39.1 & 43.6 & 60.2 & \textbf{67.5} \\
& Global MMLU Lite & 55.5 & 54.5 & 74.5 & \textbf{75.1} \\
\hline
Multi-modal & MMMU (val) & 47.3 & 48.8 & N/A & \textbf{64.9} \\
\Xhline{2.5\arrayrulewidth}
\end{tabular}
}
\begin{tablenotes}
  \scriptsize
  \item $\S$ Prior reported results
\end{tablenotes}\end{table}

\paragraph{Observed themes:}

Several key themes emerged from our analysis. While larger, more computationally expensive models generally performed better, medically-specialized models like MedGemma exhibited a distinct advantage relative to their parameter size. This size-to-performance benefit was especially pronounced in medical vision-based tasks, where smaller clinical models, in some instances, even surpassed the performance of significantly larger general-purpose models. Notably, there was a 500-fold difference in computational cost between MedGemma 4B and the most expensive comparator model, a crucial consideration for practical application development where both development and compute usage are constraints.

We also observed that performance on older, established benchmarks tended to improve with newer models. While this observation reflects genuine advancements in model capabilities, it also raises the possibility of test data leakage, as these benchmarks are publicly available and frequently used in model development and evaluation.

\section{MedGemma Fine-tuning Demonstration}
\label{sec:sft_medgemma}
\textcolor{black}{While the MedGemma models achieved strong baseline performance, users who seek to improve performance in specialized scenarios (e.g. conforming to a specific reporting style, classification with classes difficult to describe using language alone, new domains that the base models haven't been trained on) will need to further adapt the models.}

We conducted four fine-tuning experiments to demonstrate MedGemma's capacity to be adapted. There are three multimodal tasks, (1) MIMIC-CXR clinical report generation, (2) SIIM-ARC CXR pneumothorax classification \cite{siim-acr-pneumothorax-segmentation}, and (3) CRC100k histopathology patch classification \cite{Kather2018_CRC100k}, as well as one text-only task involving long-context electronic health record (EHR) question answering. In these experiments, we adapted the released models via supervised fine-tuning (SFT) for the multimodal tasks and RL for the EHR task. We note that the only task that strictly required the usage of SFT instead of RL is MIMIC-CXR report generation, in order to capture the implicit reporting style; other tasks can be fine-tuned through either SFT or RL. In our experience, the performance trade-off between these two methods needs to be established on a case-by-case basis given the dataset size, distribution shift compared to the original training datasets, and difficulty of the task.

\subsection{Fine-tuning for multimodal tasks}
\textcolor{black}{We utilized the instruction-tuned version of MedGemma 4B as the base model for the multimodal fine-tuning tasks. For MIMIC-CXR report generation tasks, the image and the indication section of the report were used as the input, and the model was trained to complete the finding and impression sections of the report. For pneumothorax classification and histopathology patch classification, the classification task was formulated as a multiple choice question, the image and a multiple choice question prompt were used as the input, and the model was trained to produce the correct choice. The model was fine-tuned through SFT which involves full parameter updates by optimizing a standard cross-entropy loss function for next token prediction. The input image and prompt were excluded from the loss computation. For SIIM-ARC and CRC100k, since these data sets provide no validation split, we reserved a random 10\% per label of the original training data to form a validation set. For each task, we conducted a hyperparameter sweep across three learning rates (1e--7, 5e--7, 1e--6) and fine-tuned the model for a single epoch. The checkpoints that achieved the highest performance on the validation set were selected for the final evaluation on the test set.}

\textcolor{black}{The results in Table~\ref{tab:downstream-sota-performance} show the effectiveness of further adapting the MedGemma models to specific use cases. The adapted models more closely approached SOTA models on these tasks \cite{zhou2024generalistlearnermultifacetedmedical,dai2024unichestconqueranddividepretrainingmultisource,chen2020bigselfsupervisedmodelsstrong}. Notably, the fine-tuned MedGemma 4B established a new SOTA performance with a RadGraph F1 score of 30.3 on the MIMIC-CXR report generation task.}

\begin{table}[t]
\centering
\caption{\textcolor{black}{\small{\textbf{MedGemma fine-tuning results.}}}}
\footnotesize
\label{tab:downstream-sota-performance}
\renewcommand{\arraystretch}{1.25}
\resizebox{\textwidth}{!}{
\begin{tabular}{l|l|l|c|c|l}
\Xhline{2.5\arrayrulewidth}
\multirow{2}{*}{\textbf{Task}} & \multirow{2}{*}{\textbf{Dataset}} & \multirow{2}{*}{\textbf{Metric}} & \multicolumn{2}{c|}{\textbf{MedGemma 4B}} & \multirow{2}{*}{\textbf{SOTA}\textsuperscript{\S}} \\
\cline{4-5}
 & & & \textbf{Out-of-box} & \textbf{Fine-tuned} & \\
\Xhline{2\arrayrulewidth}
\multirow{2}{*}{CXR report generation} & \multirow{2}{*}{MIMIC-CXR} & \multirow{2}{*}{RadGraph F1} & \multirow{2}{*}{29.5} & \multirow{2}{*}{\textbf{30.3}
} & 30.0    \\
& & & & & MedVersa \cite{zhou2024generalistlearnermultifacetedmedical} \\
\Xhline{2\arrayrulewidth}
\multirow{4}{*}{CXR binary classification} & \multirow{4}{*}{\makecell[l]{SIIM-ACR \\ Pneumothorax \\ (OOD)}} & \multirow{2}{*}{Accuracy} & \multirow{2}{*}{85.9} & \multirow{2}{*}{87.8} & \textbf{88.9} \\
& & & & & Unichest FT \cite{dai2024unichestconqueranddividepretrainingmultisource} \\
\cline{3-6}
& & \multirow{2}{*}{F1} & \multirow{2}{*}{59.7} & \multirow{2}{*}{71.5} & \textbf{72.5} \\
& & & & & Unichest FT \cite{dai2024unichestconqueranddividepretrainingmultisource} \\
\Xhline{2\arrayrulewidth}
\multirow{2}{*}{Histopathology classification} & \multirow{2}{*}{CRC100k (OOD)} & \multirow{2}{*}{Weighted F1} & \multirow{2}{*}{32.8} & \multirow{2}{*}{94.5} & \textbf{97.3} \\
& & & & & Virchow \cite{vorontsov2023virchow} \\
\Xhline{2.5\arrayrulewidth}
\end{tabular}
}
\begin{tablenotes}
  \scriptsize
  \item $\S$ Prior reported results
\end{tablenotes}
\end{table}

\subsection{Fine-tuning for EHR information retrieval and reasoning}
\label{sec:ehr}

\textcolor{black}{While numerous benchmarks assess the ability of language models to reason over EHRs, including emrQA \cite{pampari-etal-2018-emrqa}, emrKBQA \cite{raghavan-etal-2021-emrkbqa}, EHRSQL \cite{lee-etal-2024-overview}, EHRNoteQA \cite{kweon2024ehrnoteqa}, and MedAlign \cite{fleming2024medalign}, we developed a new benchmark for MedGemma specifically focused on longitudinal, outpatient EHR data. Our benchmark, which we call EHRQA, is based on a programmatic framework to generate question-answer (QA) pairs from synthetic, FHIR-formatted records. These records were produced using Synthea \cite{walonoski2018synthea}, a tool that simulates complete patient medical histories based on population statistics and disease progression models. From an initial 100 synthetic records, we selected 81 that fit within a 32k context window, splitting them into training (42), validation (20), and test (19) sets. Each patient record is comprehensive, containing hundreds to thousands of FHIR entries across various resource types like Conditions, Medications, and Observations. A key limitation of this dataset, however, is the absence of clinical notes, which we hope to address in future work.}

\textcolor{black}{Our QA generation framework treated each patient's FHIR record as a collection of ground truth facts. It expanded upon these facts through a multi-hop reasoning process \cite{yang-etal-2018-hotpotqa} that integrated information from external medical knowledge bases (SNOMED, RxNorm, LOINC), applied programmatic reasoning (e.g., temporal, arithmetic), and leveraged inter-dependencies within the patient's data programmatically. These derived facts are then converted into natural language QA pairs using templated questions and large language model (LLM) rephrasing. In collaboration with clinicians, we developed 42 distinct question types, grouped into 10 categories, to reflect queries from both healthcare professionals and consumers (see Appendix Table~\ref{tab:ehrqa_examples} for more details). This process generated approximately 200 questions per patient. In total, there are 10,437 QA examples in the training split, 5,133 QA examples in the validation split, and 4,377 QA examples in the test split. All questions were designed for automated evaluation (e.g., regex matching) and support multiple-choice, Yes/No, and simple free-response formats. For evaluation, the model is prompted with a condensed, plain-text representation of the patient's FHIR data and the corresponding question. Final accuracies are calculated by averaging the scores across the 10 question categories. The prompts used for rephrasing and evaluation are available in Appendix Table~\ref{tab:ehrqa_prompts}.}

\textcolor{black}{Our initial results revealed a performance gap between MedGemma 27B and larger models. To address this, we fine-tuned MedGemma 27B using RL on the EHRQA training set. Applying RL to the MedGemma 27B resulted in a substantial improvement in its EHRQA accuracy, closing the gap with top-performing models, as shown in Table~\ref{tab:ehrqa_results}. Notably, the greatest gains were observed in question categories requiring reasoning across inter-dependent records (Appendix Figure~\ref{fig:ehrqa_results_figure}). This experiment underscores the potential of fine-tuning smaller, specialized models like MedGemma for effective use in EHR applications.}

\begin{table}[h]
\centering
\caption{\textcolor{black}{\small{\textbf{EHRQA accuracy results (OOD)}}}}
\label{tab:ehrqa_results}
\resizebox{\textwidth}{!}{
\begin{tabular}{l|c|c|c|c|c|c}
\Xhline{2.5\arrayrulewidth}
\multicolumn{1}{l|}{} & \multicolumn{3}{c|}{\textbf{Small Models}} & \multicolumn{3}{c}{\textbf{Large Models}} \\
\cmidrule(lr){2-7}
\textbf{Type} & \makecell[c]{\textbf{MedGemma 27B }\\ \textbf{Text-only}} & \textbf{Gemma 3 27B}  & \makecell[c]{\textbf{MedGemma 27B} \\ \textbf{Text-only (RL)}} & \textbf{Gemini 2.5 Flash} & \textbf{Gemini 2.5 Pro} & \textbf{o3}\\
\Xhline{2.5\arrayrulewidth}
{\textbf{EHRQA}} & 86.3 & 84.2 & \textbf{93.6} & 95.0 & \textbf{95.4} & 92.5 \\
\Xhline{2.5\arrayrulewidth}
\end{tabular}
}
\end{table}

\section{MedSigLIP Evaluations}
\label{sec:eval_medsiglip}
MedSigLIP was evaluated both on zero-shot classification performance and linear probe classification performance (also referred to as data-efficient classification). Zero-shot classification performance was assessed as a measure of baseline performance, and linear probe performance was assessed to gauge performance on target conditions after additional training with logistic regression.

Evaluation datasets spanned four modalities: chest X-ray, dermatology, ophthalmology and histopathology. For chest X-ray, we followed the evaluation datasets and framework in ELIXR~\cite{xu2023elixr}. For linear probing, we utilized the same data from CheXpert and CXR14 and evaluated on seven findings (atelectasis, cardiomegaly, airspace opacity,
fracture, pneumothorax, consolidation, pleural effusion, and pulmonary edema). For zero-shot, we compared with ELIXR on the 13 positive findings from CheXpert test set.
For dermatology, we evaluated on US-Derm MCQA. The US-Derm MCQA test set has 79 dermatological conditions that overlap with the classification labels from the training dataset, thus we pick this 79-condition subset (1612 patients) for our vision based classification and zero-shot evaluation. For ophthalmology, we used the EyePACS test dataset. For histopathology, patches were extracted from whole slide images across a variety of tissue types and tasks as in~\citet{yang2024advancing} and these patches were treated as individual images. See Appendix Table~\ref{tab:med_siglip_classes} for a description of the detailed classes for the evaluations.

\paragraph{Zero-shot evaluation approach and metrics:} For each class within each condition, one or more text prompts were used to represent the class. When there were multiple prompts for the same class, the text embeddings for each prompt were averaged together to obtain a single embedding per class. Cosine similarity was then calculated between the image embedding and text embeddings for the class options, softmax was applied to output scores, and the AUC (area under receiver operating characteristic curve) was calculated. In the case of multiclass conditions, 1-vs-all AUC was calculated. For dermatology, zero-shot prompts are directly their condition names. Prompts for ophthalmology, chest X-ray, histopathology are provided in Appendix Tables~\ref{tab:zero_shot_ophthalmology_prompts},~\ref{tab:medsiglip_cxr_prompts}, and~\ref{tab:medsiglip_path_prompts}.

\paragraph{Linear probe evaluation approach and metrics:} Linear probe (data efficient) classification was assessed by extracting image embeddings from MedSigLIP (without use of the text encoder) and training a logistic regression using the SAGA solver~\cite{defazio2014saga} on the train set embeddings, with the hyper-parameters chosen on the validation set, then evaluating on the test set.

\section{MedSigLIP Results}
\label{sec:results_medsiglip}
Zero-shot and linear probe classification results and evaluation tasks for MedSigLIP and corresponding Health AI Developer Foundations (HAI-DEF) \cite{kiraly2024health} models are summarized below. For the HAI-DEF Derm Foundation and Path Foundation models, zero-shot classification was not possible with the comparator models because of their image-only nature.

\begin{figure}[htp]
    \centering
    \includegraphics[width=0.7\linewidth]{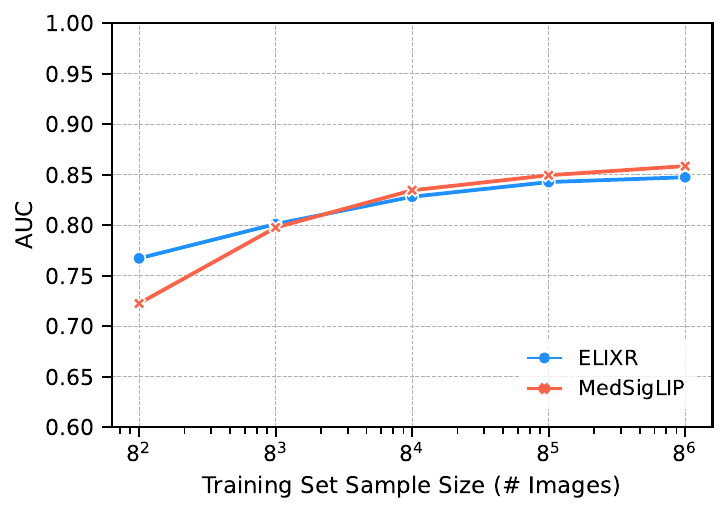}
    \caption{\small{\textbf{Average results for data efficient learning on 7 Chest X-ray findings on CheXpert and CXR14 datasets compared to HAI-DEF's CXR Foundation model based on ELIXR~\cite{xu2023elixr}. 
    Individual results per condition per dataset can be found in Appendix Figures~\ref{fig:cxr_chexpert_metric} and~\ref{fig:cxr_cxr14_metric}.
    }}}
    \label{fig:avg_metric}
\end{figure}

On CXR, MedSigLIP is compared to the HAI-DEF CXR Foundation model, which is based on ELIXR \cite{xu2023elixr}, with zero-shot results shown in \cref{tab:med_siglip_cxr_results} and linear probe classification results shown in Figure~\ref{fig:avg_metric}. On average, MedSigLIP's zero-shot CXR performance was 2.0\% higher than CXR Foundation despite MedSigLIP's lower image resolution (448$\times$448 vs 1280$\times$1280) and multi-domain expertise, suggesting that MedSigLIP can serve as a strong foundation model. Notably, classification of fractures, which has historically proven difficult, improved over ELIXR by 7.1\%. Across the 7 findings used for linear probing, MedSigLIP also demonstrated strong performance when training set sample size is larger or equal to 512 examples, Figure~\ref{fig:avg_metric}.

\begin{table}[htbp]
 \centering
 \footnotesize
 \caption{\small\textbf{AUCs for dermatology, ophthalmology, and histopathology findings with MedSigLIP and HAI-DEF image models~\cite{kiraly2024health}.}}
 \label{tab:med_siglip_results}
 \resizebox{\textwidth}{!}{
 \begin{tabular}{@{}ll|cc|ccc|ccc@{}}
 \toprule
 Domain & Finding & N & No. & \multicolumn{3}{c|}{MedSigLIP} & \multicolumn{2}{c}{HAI-DEF (image)\textsuperscript{\S}} \\
 & & & Classes & Resolution & Zero-Shot & Linear Probe & Resolution & Linear Probe \\
 \midrule
 Dermatology & Skin Conditions & 1612 & 79 & 448x448 & 0.851 & \textbf{0.881} & 448$\times$448  & 0.843 \\
 \midrule
 Ophthalmology & Diabetic Retinopathy & 3161 & 5 & 448x448 & 0.759 & \textbf{0.857} & N/A & N/A \\
 \midrule
 \multirow{8}{*}{Histopathology}
 & Invasive Breast Cancer & 5000 & 3 & \multirow{8}{*}{448$\times$448} & 0.933 & 0.930 & \multirow{8}{*}{224$\times$224} & 0.943 \\
 & Breast NP & 5000 & 3 & & 0.721 & 0.727 &  & 0.758 \\
 & Breast TF & 5000 & 3 & & 0.780 & 0.790 &  & 0.832 \\
 & Cervical Dysplasia & 5000 & 3 & & 0.889 & 0.864 &  & 0.898 \\
 & Prostate Cancer Needle Core Biopsy & 5000 & 4 & & 0.892 & 0.886 &  & 0.915 \\
 & Radical Prostatectomy & 5000 & 4 & & 0.896 & 0.887 &  & 0.921 \\
 & TCGA Study Types & 5000 & 10 & & 0.922 & 0.970 &  & 0.964 \\
 & Tissue Types & 5000 & 16 & & 0.930 & 0.972 &  & 0.947 \\
 \cline{5-9} 
 \multicolumn{2}{@{}l|}{\textbf{Average}} & & & & 0.870 & 0.878 & & \textbf{0.897} \\
 \bottomrule
 \end{tabular}
 }
\begin{tablenotes}
  \scriptsize
  \item $\S$ Prior reported results. HAI-DEF (image) refers to the image-only foundation models in HAI-DEF: Derm Foundation and Path Foundation models while MedSigLIP is a single model.
\end{tablenotes}
\end{table}

Dermatology, ophthalmology and histopathology results are summarized in \cref{tab:med_siglip_results}. For dermatology, both MedSigLIP zero-shot and linear probes outperformed linear probes with Derm Foundation on the task of distinguishing between 79 skin conditions. For ophthalmology, on the 5-class task of classifying diabetic retinopathy (none, mild, moderate, severe, and proliferative), linear probe performance exceeded zero-shot performance by 9.8\%, though no HAI-DEF model currently exists for comparison. For histopathology, linear probing achieved a modest improvement over zero-shot classification using MedSigLIP, and performing close to Path Foundation which has linear probe performance that is 1.9\% better on average.

\begin{table}[htbp]
 \centering
 \footnotesize
 \caption{\small\textbf{Zero-shot AUCs for chest X-ray findings with MedSigLIP and HAI-DEF's chest X-ray foundation model based on ELIXR~\cite{xu2023elixr}.}}
 \label{tab:med_siglip_cxr_results}
 \resizebox{\textwidth}{!}{
 \begin{tabular}{@{}ll|cc|cc|cc@{}}
 \toprule
 Domain & Finding & N & No. & \multicolumn{2}{c|}{MedSigLIP} & \multicolumn{2}{c}{HAI-DEF (image)\textsuperscript{\S}} \\
 & & & Classes & Resolution & Zero-Shot & Resolution & Zero-Shot \\
 \midrule
 \multirow{13}{*}{CXR}
 & Enlarged Cardiomediastinum & \multirow{13}{*}{518} & \multirow{13}{*}{2} & \multirow{13}{*}{448$\times$448} & 0.858  & \multirow{13}{*}{1280$\times$1280} & 0.800 \\
 & Cardiomegaly & & & & 0.904 & & 0.891 \\
 & Lung Opacity & & & & 0.931 & & 0.888  \\
 & Lung Lesion & & & & 0.822  & & 0.747  \\
 & Consolidation & & & & 0.880  & & 0.875  \\
 & Edema & & & & 0.891  & & 0.880  \\
 & Pneumonia & & & & 0.864  & & 0.881  \\
 & Atelectasis & & & & 0.836  & & 0.754  \\
 & Pneumothorax & & & & 0.862 & & 0.800  \\
 & Pleural Effusion & & & & 0.914  & & 0.930  \\
 & Pleural Other & & & & 0.650 &  & 0.729  \\
 & Fracture & & & & 0.708  & & 0.637  \\
 & Support Devices & & & & 0.852  & & 0.894  \\
 \cline{5-8} 
 \multicolumn{2}{@{}l|}{\textbf{Average}} & & & & \textbf{0.844}  & & 0.824  \\
 \bottomrule
 \end{tabular}
 }
 \begin{tablenotes}
  \scriptsize
  \item $\S$ Prior reported results
  \end{tablenotes}
\end{table}

MedSigLIP offers strong baseline performance across a variety of medical domains as a single model and generally does well with zero-shot classification even when compared to linear probes with dedicated, domain-specific models. Further training with
task-specific images and logistic regression can yield stronger results.

\section{Discussion}
\label{sec:discussion}
We introduced MedGemma, a new collection of medical vision-language foundation models and MedSigLIP, a multi-domain medical image encoder. These models were built upon Gemma 3, with optimization for medical domains. We evaluated across a range of medical benchmarks across clinical reasoning, biomedical knowledge, report generation, and medical image classification, finding strong performance for MedGemma and MedSigLIP. Performance improved further after fine-tuning, highlighting the potential for these open models to be used as a starting point for developing useful AI applications for healthcare.

With an increasing number of options available to developers building AI applications in healthcare, MedGemma provides specific advantages over general models. These advantages are largely due to optimized incorporation of domain specific data for both pre-training and post-training and are illustrated by the improvements over base Gemma 3 models across all benchmarks evaluated and the achievement of performance on par with much larger models.

When compared to general API-based models like Gemini, MedGemma is likely the preferred model if the use case requires any of the following: a frozen model for documentation and reliability, sensitivity to training or inference costs, ability to run locally or offline, specific medical image and text capabilities, or full control over model adaptation. Large models like Gemini remain a viable choice where the user requires optimal broad performance without the above constraints, and large models may additionally be used in concert with models like MedGemma in agentic settings.

The MedGemma collection of models enables a wide range of potential downstream applications for the developer community. The multimodal capabilities, including access to image and text embeddings, may be particularly useful for medical image retrieval. This could aid in interpretation by referencing similar past cases as well as enabling development of research cohorts and creating educational tools. MedGemma allows for the integration of diverse data, linking radiology, histopathology, ophthalmology, and dermatology images with clinical information. The specialized text capabilities of the models can also extract key concepts from imaging reports and clinical notes, streamlining tasks such as matching patients for clinical trials, conducting pharmacovigilance reviews, or analyzing healthcare quality metrics. The models' ability to understand medical images and generate reports can also be fine-tuned to better assist radiologists and other clinicians in their workflow and improve how findings are communicated to patients. In addition to standalone use, these models can also serve as powerful tools within agentic frameworks, combining abilities across different modalities for customized and comprehensive solutions.

In this report, we evaluated the performance of MedGemma and MedSigLIP on a broad set of established benchmarks in order to provide a snapshot of the model capabilities. However, we note that limitations exist for these benchmarks. For one, automated benchmarks represent only the first step towards validating real-world utility \cite{alaa2025medical, mahmood2025benchmarking}. Additionally, some benchmarks may be near saturation in terms of model performance, with minimal headroom for improvements, thus hindering the measurement of progress. As such, further work is warranted to continue evaluation of these models on new, high quality (and more challenging) benchmarks aimed at better reflecting real-world utility \citep[e.g.][]{bedi2025medhelm}. More work is also needed to understand the performance capabilities and requirements in regard to actual application development, including their incorporation into agentic frameworks. These efforts will inform optimal use cases as well as development of future model versions with extended capabilities.

We openly released MedGemma and MedSigLIP to facilitate their widespread evaluation, improvement, and adaptation by the community. Openness is critical in many healthcare applications, as it provides developers with predictability and the flexibility for extensive model adaptation and evaluation. We hope that our approach accelerates the development of AI applications across a broad array of healthcare use cases.

\section{Conclusion}
\label{sec:conclusion}
In this work, we showed that MedGemma models demonstrate robust capabilities across a variety of vision-language and text-only medical tasks. We also showed that MedSigLIP demonstrates robust multi-domain capabilities, and can thus serve as a strong medical foundation model. The breadth and efficiency of these models offers exciting possibilities to address a range of use cases. At the same time, thoughtful validation of safety, performance, and reliability for any downstream applications remains a critical aspect to advance the use of multimodal AI models in medicine. By providing these MedGemma and MedSigLIP models to the developer community with a permissive license, we hope to see them enable useful and innovative medical applications. 

\section{Model availability}
The models have been released openly at the main Google Health AI Developer Foundations site at \mbox{\url{https://goo.gle/hai-def}}. Further details specifically about the MedGemma collection of models can be found at \mbox{\url{https://goo.gle/medgemma}}.

\newpage
\section{Contributions and Acknowledgments}
\subsubsection*{Contributions}

\begin{multicols*}{2}
\setlength{\columnsep}{5pt}
\small
\vspace{-1.0\baselineskip}
\begin{itemize}[leftmargin=1em,rightmargin=0em]
\setlength\itemsep{0pt}
    \item[] \textbf{Technical Leads} \\\vspace{-6pt}
    \item[] Lin Yang\textsuperscript{$\dagger$}
    \item[] Andrew Sellergren\textsuperscript{*}
    \item[] Sahar Kazemzadeh\textsuperscript{*}
    \item[] Fereshteh Mahvar
\end{itemize}

\vspace{-1.0\baselineskip}
\begin{itemize}[leftmargin=1em,rightmargin=0em]
\setlength\itemsep{0pt}
    \item[] \textbf{Core contributors} \\\vspace{-6pt}
    \item[] Tiam Jaroensri
    \item[] Atilla Kiraly
    \item[] Madeleine Traverse
    \item[] Timo Kohlberger
    \item[] Shawn Xu
    \item[] Fayaz Jamil
    \item[] Cían Hughes
    \item[] Charles Lau
    \item[] Justin Chen
    \item[] Liron Yatziv
    \item[] Tiffany Chen
    \item[] Bram Sterling
    \item[] Stefanie Anna Baby
    \item[] Susanna Maria Baby
    \item[] Jeremy Lai
    \item[] Samuel Schmidgall
    \item[] Lu Yang
    \item[] Kejia Chen
    \item[] Per Bjornsson
    \item[] Shashir Reddy
    \item[] Ryan Brush
    \item[] Kenneth Philbrick
    \item[] Mercy Asiedu
    \item[] Ines Mezerreg
    \item[] Howard Hu
    \item[] Howard Yang
    \item[] Richa Tiwari
    \item[] Sunny Jansen
    \item[] Preeti Singh
    \item[] Yun Liu
    \item[] Shekoofeh Azizi
\end{itemize}

\vspace{-1.0\baselineskip}
\begin{itemize}[leftmargin=1em,rightmargin=0em]
\setlength\itemsep{0pt}
    \item[] \textbf{Contributors} \\\vspace{-6pt}
    \item[] Aishwarya Kamath
    \item[] Johan Ferret
    \item[] Shreya Pathak
    \item[] Nino Vieillard
    \item[] Ramona Merhej
    \item[] Sarah Perrin
    \item[] Tatiana Matejovicova
    \item[] Alexandre Ramé
    \item[] Morgane Riviere
    \item[] Louis Rouillard
    \item[] Thomas Mesnard
    \item[] Geoffrey Cideron
    \item[] Jean-bastien Grill
    \item[] Sabela Ramos
    \item[] Edouard Yvinec
    \item[] Michelle Casbon
    \item[] Elena Buchatskaya
    \item[] Jean-Baptiste Alayrac
    \item[] Dmitry (Dima) Lepikhin
    \item[] Vlad Feinberg
    \item[] Sebastian Borgeaud
    \item[] Alek Andreev
    \item[] Cassidy Hardin
    \item[] Robert Dadashi
    \item[] Léonard Hussenot
    \item[] Armand Joulin
    \item[] Olivier Bachem
\end{itemize}

\vspace{-1.0\baselineskip}
\begin{itemize}[leftmargin=1em,rightmargin=0em]
\setlength\itemsep{0pt}
    \item[] \textbf{Sponsors} \\\vspace{-6pt}
    \item[] Yossi Matias
    \item[] Katherine Chou
    \item[] Avinatan Hassidim
    \item[] Kavi Goel
    \item[] Clement Farabet
    \item[] Joelle Barral
    \item[] Tris Warkentin
    \item[] Jonathon Shlens
    \item[] David Fleet
\end{itemize}

\vspace{-1.0\baselineskip}
\begin{itemize}[leftmargin=1em,rightmargin=0em]
\setlength\itemsep{0pt}
    \item[] \textbf{Launch Support} \\\vspace{-6pt}
    \item[] Victor Cotruta
    \item[] Omar Sanseviero
    \item[] Gus Martins
    \item[] Phoebe Kirk
    \item[] Anand Rao
\end{itemize}

\vspace{-1.0\baselineskip}
\begin{itemize}[leftmargin=1em,rightmargin=0em]
\setlength\itemsep{0pt}
    \item[] \textbf{Leads} \\\vspace{-6pt}
    \item[] Shravya Shetty
    \item[] David F. Steiner
    \item[] Can Kirmizibayrak
    \item[] Rory Pilgrim\textsuperscript{$\dagger$}
    \item[] Daniel Golden\textsuperscript{$\dagger$}
    \item[] \footnotesize{$*$ Co-first author}
    \item[] \footnotesize{$\dagger$ Co-last author}
\end{itemize}
\end{multicols*}

\subsubsection*{Acknowledgements}
Many teams from both Google Research and Google DeepMind collaborated extensively on this project.
We thank Ellery Wulczyn and Dale Webster for their feedback and insight, which significantly enhanced this report.
We thank Naama Hammel, Liam Foster, and Kapil Parakh for their review of the qualitative examples shown in this manuscript. We thank Andreas Steiner and Xiao Wang for sharing their expertise with the Big Vision infrastructure. The results shown here are in part based upon data generated by the \href{https://www.cancer.gov/tcga}{TCGA Research Network}.  

\subsubsection*{Use of AI in Manuscript Preparation}
The introduction section of this manuscript was drafted manually and then further refined using Gemini 2.5 Pro. Final manual checks were performed to ensure content accuracy. The authors take full responsibility for the content.

\clearpage
\bibliography{main}

\newpage
\appendix
\label{sec:appendix}
\setcounter{section}{0}
\setcounter{figure}{0}
\setcounter{table}{0}
\setcounter{equation}{0}

\noindent \textbf{\LARGE{Appendix}}\\
\normalfont

\renewcommand{\thefigure}{A\arabic{figure}} 
\renewcommand{\thetable}{A\arabic{table}}

\section{Manual Evaluation of Radiology reports}

\cref{tab:human_report_rubric} provides the detailed rubric definitions provided to the radiologist for scoring the generated CXR reports. 

\begin{table}[htp]
\small
\centering
\caption{\small{\textbf{Human evaluation rubric comparing AI generated radiology reports to original reports.}}}
\label{tab:human_report_rubric}
\definecolor{A2}{RGB}{87,172,255}
\definecolor{A1}{RGB}{194,226,237}
\definecolor{Csame}{RGB}{222,222,222}
\definecolor{B1}{RGB}{255,221,227}
\definecolor{B2}{RGB}{255,138,117}
\definecolor{Xneither}{RGB}{135,135,135}

\begin{tabular}{p{3.3cm}p{13cm}} 
\toprule
\textbf{Rubric Score} & \textbf{Rubric Definition}  \\
\hline
\\
\hline \rowcolor{A2}
\multirow{2}{*}{{\textbf{AI >> Original Report}}}  &   \small{AI report captures key clinically relevant findings that are not found in original report. AI report would result in correct patient management and original report would not.}        \\ \hline \rowcolor{A1}
\multirow{2}{*}{\textbf{AI > Original Report}}  & AI report captures more relevant findings, but both would result in the same correct patient management.   \\ \hline \rowcolor{Csame}
\multirow{2}{*}{\textbf{AI $\sim$ Original Report}}   & \RaggedRight{Both reports capture similar findings in the image and would result in correct patient management.}  \\ \hline \rowcolor{B1}
\multirow{2}{*}{\textbf{AI < Original Report}}  & Original report captures more relevant findings, but both would result in the same correct patient management. \\ \hline \rowcolor{B2}
\multirow{2}{*}{\textbf{AI << Original Report}}  & Original report captures key clinically relevant findings that are not found in AI report. original report would result in correct patient management and AI report would not.  \\ \hline  \rowcolor{Xneither}
\multirow{1}{*}{\textbf{X}}   & Neither report would result in correct patient management.  \\
\bottomrule
\end{tabular}
\end{table}

\clearpage

\section{Evaluation Prompts}

For evaluating MedSigLIP zero-shot tasks, Tabes~\ref{tab:med_siglip_classes} to \ref{tab:medsiglip_path_prompts} list the prompts used.

For evaluating MedGemma, the default temperature of 0.0 was used. All other models made use of their default system temperature. Two different system prompts were used in providing results for other LLMs directly evaluated:

\begin{itemize}

\item \textit{You are a helpful radiology assistant.} For radiology-based tasks.

\item \textit{You are a helpful medical assistant.} For all other medical tasks.
    
\end{itemize}

Per-dataset prompts used are shown in Appendix Tables~\ref{tab:text_eval_prompts},~\ref{tab:classification_eval_prompts}.

\begin{table}[htbp]
\centering
\footnotesize
\caption{\small\textbf{MedSigLIP Class Definitions}}
\caption*{\small Classes for the findings that MedSigLIP was evaluated on.}
\label{tab:med_siglip_classes}
\begin{tabularx}{\textwidth}{@{}l l c X@{}}
\toprule
Domain & Finding & Num Classes & Class Names \\
\midrule
\multirow{1}{*}{Radiology}
& CXR & 2 & Positive/negative for Cardiomegaly, Consolidation, Edema, Enlarged cardiomediastinum, Fracture, Lung lesion, Lung opacity, Pleural effusion, Pleural other, Pneumonia, Pneumothorax,Support devices.\\
\midrule
\multirow{8}{*}{Histopathology}
& Invasive Breast Cancer & 3 & Benign, Invasive Carcinoma, DCIS \\
& Breast NP  & 3 & NP1, NP2, NP3 \\
& Breast TF  & 3 & TF1, TF2, TF3 \\
& Cervical Dysplasia  & 3 & Normal, CIN Grade 1, CIN Grade 2+ \\
& Prostate Cancer Needle Core Biopsy & 4 & Benign, GP3, GP4, GP5 \\
& Radical Prostatectomy  & 4 & Benign, GP3, GP4, GP5 \\
& TCGA Study Types & 10 & BLCA, BRCA, COAD, HNSC, KIRC, LIHC, LUAD, LUSC, OV, STAD \\
& Tissue Types & 16 & Appendix, Breast, Cervix, Colon, Fallopian Tube, Gallbladder, Liver, Lymph node, Ovary, Placenta, Prostate, Skin, Thyroid, Upper GI, Uterus, Vas deferens \\
\midrule
Dermatology
& Skin Conditions & 79 & \tiny Acanthosis nigricans, Acne, Acne keloidalis, Actinic Keratosis, Allergic Contact Dermatitis, Alopecia Areata, Amyloidosis of skin, Androgenetic Alopecia, Angiokeratoma of skin, Atypical Nevus, Basal Cell Carcinoma, Bullous Pemphigoid, Burn of skin, Candida, Clavus, Comedone, Condyloma acuminatum, Cutaneous T Cell Lymphoma, Cutaneous lupus, Cutaneous sarcoidosis, Cyst, Dermatofibroma, Drug Rash, Eczema, Erythema ab igne, Erythema multiforme, Folliculitis, Folliculitis decalvans, Fordyce spots, Granuloma annulare, Hemangioma, Herpes Zoster, Hidradenitis, Idiopathic guttate hypomelanosis, Infected skin lesions, Inflicted skin lesions, Insect Bite, Intertrigo, Irritant Contact Dermatitis, Keratosis pilaris, Knuckle pads, Lentigo, Lichen Simplex Chronicus, Lichen planopilaris, Lichen planus/lichenoid eruption, Lichen sclerosus, Lipodermatosclerosis, Lipoma, Livedo reticularis, Melanocytic Nevus, Melanoma, Melasma, Milia, Nevus sebaceous, O/E - ecchymoses present, Onychomycosis, Perioral Dermatitis, Photodermatitis, Pigmented purpuric eruption, Pityriasis lichenoides, Pityriasis rosea, Post-Inflammatory hyperpigmentation, Prurigo nodularis, Psoriasis, Pyoderma Gangrenosum, Pyogenic granuloma, Rosacea, SCC/SCCIS, SK/ISK, Scabies, Scar Condition, Seborrheic Dermatitis, Skin Tag, Stasis Dermatitis, Tinea, Tinea Versicolor, Urticaria, Verruca vulgaris, Vitiligo \\
\midrule
Ophthalmology
& Diabetic Retinopathy (DR) & 5 & No DR, Mild DR, Moderate DR, Severe DR, Proliferative DR \\
\bottomrule
\end{tabularx}

\end{table}

\begin{center}
\begin{table}[htb]
\centering
\caption{Zero shot prompts for ophthalmology}\label{tab:zero_shot_ophthalmology_prompts}
\begin{tabular}{@{}ll@{}}
\toprule
\multicolumn{2}{c}{\textit{Ophthalmology}} \\
\midrule
\textbf{Finding} & \textbf{Severity} \\
\midrule
Diabetic retinopathy & \makecell[l]{%
  diabetic retinopathy severity: none. \\%
  diabetic retinopathy severity: mild. \\%
  diabetic retinopathy severity: moderate. \\%
  diabetic retinopathy severity: severe. \\%
  diabetic retinopathy severity: proliferative.%
} \\
\bottomrule
\end{tabular}
\end{table}
\end{center}

\begin{table}[htb]
    \centering
    \caption{Zero shot prompts for CXR}\label{tab:medsiglip_cxr_prompts}
    \resizebox{\textwidth}{!}{
    \begin{tabular}{@{}lll@{}}
        \toprule
        \multicolumn{3}{c}{\textit{Chest X-ray}} \\
        \midrule
        \textbf{Finding} & \textbf{Condition Absent} & \textbf{Condition Present} \\
        \midrule
        Atelectasis & \makecell[l]{no atelectasis \\ no acute cardiopulmonary process} & \makecell[l]{adjacent atelectasis \\ bibasilar atelectasis \\ there is atelectasis} \\
        \midrule
        Cardiomegaly & \makecell[l]{heart size is normal \\ cardiac size is within normal limits \\ cardiothoracic ratio within normal limits \\ no acute cardiopulmonary process \\ normal study \\ no evidence of cardiomegaly} & \makecell[l]{mild cardiomegaly \\ moderate cardiomegaly \\ severe cardiomegaly \\ enlarged cardiac silhouette} \\
        \midrule
        Consolidation & \makecell[l]{no focal consolidation \\ no acute cardiopulmonary process \\ normal study} & \makecell[l]{alveolar consolidation \\ densely consolidated \\ lobe consolidation \\ airspace consolidation \\ bibasilar consolidations \\ suggestive of consolidation} \\
        \midrule
        Edema & \makecell[l]{no pulmonary edema \\ no acute cardiopulmonary process \\ normal study} & \makecell[l]{mild pulmonary edema \\ moderate pulmonary edema \\ severe pulmonary edema} \\
        \midrule
        Enlarged cardiomediastinum & \makecell[l]{no acute cardiopulmonary process \\ cardiomediastinal silhouette is normal} & widened cardiomediastinum \\
        \midrule
        Fracture & \makecell[l]{no acute cardiopulmonary process \\ normal study} & \makecell[l]{rib fractures \\ rib fracture} \\
        \midrule
        Lung lesion & no acute cardiopulmonary process & \makecell[l]{lytic lesion \\ cavitary lesion \\ parenchymal lesion} \\
        \midrule
        Lung opacity & \makecell[l]{no focal opacity \\ lung volumes are normal \\ normal study \\ normal lung volumes \\ lungs are clear \\ no focal consolidation \\ no evidence of airspace consolidation \\ no infiltrate \\ no airspace opacity \\ pulmonary parenchyma is clear} & \makecell[l]{bilateral opacities \\ basal opacity \\ opacification is present \\ bibasilar opacities \\ increased opacification \\ consolidative opacity \\ parenchymal opacities \\ airspace opacification cannot be excluded \\ there is airspace opacification} \\
        \midrule
        Pleural effusion & \makecell[l]{no evidence of pleural effusion \\ no acute cardiopulmonary process \\ normal study} & \makecell[l]{left pleural effusion \\ right pleural effusion \\ bilateral pleural effusions} \\
        \midrule
        Pleural other & no acute cardiopulmonary process & \makecell[l]{blunting of costophrenic angle \\ pleural thickening} \\
        \midrule
        Pneumonia & \makecell[l]{lungs are clear \\ no acute cardiopulmonary process} & suggestive of pneumonia \\
        \midrule
        Pneumothorax & \makecell[l]{no pneumothorax \\ no acute cardiopulmonary process \\ normal study} & apical pneumothorax \\
        \midrule
        Support devices & no acute cardiopulmonary process & \makecell[l]{monitoring and support devices \\ NG tube \\ ET tube \\ catheter \\ PIC line} \\
        \bottomrule
    \end{tabular}}
\end{table}

\FloatBarrier

\begin{longtable}{@{}p{4cm}p{1cm}p{9cm}@{}}
  \caption{Zero shot prompts for histopathology tasks}
  \label{tab:medsiglip_path_prompts}\\
  \toprule
  \textbf{Condition} & \textbf{Class} & \textbf{Prompt} \\
  \midrule
  \endfirsthead

  \multicolumn{3}{@{}l}{\textit{(Continued)}} \\
  \toprule
  \textbf{Condition} & \textbf{Class} & \textbf{Prompt} \\
  \midrule
  \endhead

  \midrule
  \multicolumn{3}{r@{}}{\textit{(Continued on next page)}} \\
  \endfoot

  \bottomrule
  \endlastfoot

  \multicolumn{3}{@{}l}{\textbf{Invasive Breast Cancer}} \\

  & 1 &
    \begin{itemize}[noitemsep,topsep=0pt,leftmargin=*]
      \item region of an HE histopathology image showing benign breast tissue
      \item HE‐stained image demonstrating normal breast tissue architecture
      \item histopathology image with breast lobules and ducts consistent with benign breast tissue
      \item microscopic view of breast tissue with regular uniformly spaced glands indicative of benign breast tissue
      \item HE histopathology image showing normal breast stroma and absence of cellular atypia characteristic of benign breast tissue
    \end{itemize} \\

  & 2 &
    \begin{itemize}[noitemsep,topsep=0pt,leftmargin=*]
      \item region of an HE histopathology image showing invasive breast carcinoma
      \item HE‐stained region demonstrating features of invasive breast carcinoma
      \item histopathology image with areas of dense cellularity and atypical cells consistent with invasive breast carcinoma
      \item microscopic view of breast tissue with disorganized glandular architecture consistent with invasive carcinoma
      \item HE histopathology image with regions of stromal invasion by malignant cells indicating invasive breast carcinoma
    \end{itemize} \\

  & 3 &
    \begin{itemize}[noitemsep,topsep=0pt,leftmargin=*]
      \item region of an HE histopathology image showing ductal carcinoma in situ (DCIS)
      \item HE‐stained image demonstrating abnormal cells confined within the breast duct
      \item histopathology image: atypical cells within the duct consistent with DCIS
      \item microscopic view of breast tissue showing abnormal cellular proliferation within a duct consistent with DCIS
      \item HE histopathology section with non-invasive intraductal malignant cells suggestive of DCIS
    \end{itemize} \\

  \addlinespace
  \multicolumn{3}{@{}l}{\textbf{Breast NP}} \\

  & 1 &
    \begin{itemize}[noitemsep,topsep=0pt,leftmargin=*]
      \item HE histopathology image demonstrating invasive breast carcinoma with low-grade nuclear features (nucleopleomorphism score 1)
      \item region of invasive breast carcinoma exhibiting relatively uniform nuclei with inconspicuous nucleoli consistent with a low nucleopleomorphism score
      \item microscopic view of invasive breast carcinoma showing limited variation in nuclear shape and size (nucleopleomorphism score 1)
      \item focus on infiltrating tumor cells within an HE‐stained image demonstrating bland nuclear features suggestive of low-grade invasive breast carcinoma
      \item HE histopathology section of invasive breast carcinoma with well-formed glands and minimal nuclear atypia indicating a low nucleopleomorphism score
    \end{itemize} \\

  & 2 &
    \begin{itemize}[noitemsep,topsep=0pt,leftmargin=*]
      \item HE histopathology image demonstrating invasive breast carcinoma with moderate nuclear pleomorphism (score 2)
      \item region of invasive breast carcinoma exhibiting some variation in nuclear size and shape with some visible nucleoli (nucleopleomorphism score 2)
      \item microscopic view of invasive breast carcinoma showing moderate nuclear atypia including enlarged nuclei and prominent nucleoli (nucleopleomorphism score 2)
      \item focus on tumor cells with increased nuclear irregularity compared to normal breast tissue consistent with a nucleopleomorphism score of 2
      \item HE histopathology section of invasive breast carcinoma with moderately pleomorphic nuclei and discernible nucleoli (nucleopleomorphism score 2)
    \end{itemize} \\

  & 3 &
    \begin{itemize}[noitemsep,topsep=0pt,leftmargin=*]
      \item HE histopathology image demonstrating invasive breast carcinoma with high-grade nuclear features (nucleopleomorphism score 3)
      \item microscopic view of invasive breast carcinoma showing significant nuclear atypia including marked variation in size, shape, and prominent nucleoli (nucleopleomorphism score 3)
      \item region of invasive breast carcinoma exhibiting large nuclei with considerable variation in size and shape as well as large irregular nucleoli (nucleopleomorphism score 3)
      \item HE‐stained image highlighting markedly pleomorphic nuclei of tumor cells consistent with high-grade invasive breast carcinoma
      \item focus on an area of invasive breast carcinoma demonstrating significant nuclear abnormalities and variability consistent with a nucleopleomorphism score of 3
    \end{itemize} \\

  \addlinespace
  \multicolumn{3}{@{}l}{\textbf{Breast TF}} \\

  & 1 &
    \begin{itemize}[noitemsep,topsep=0pt,leftmargin=*]
      \item well-formed tubules are visible within this region of invasive breast carcinoma on an HE histopathology image (tubule formation score 1)
      \item this microscopic view demonstrates a predominance of well-defined glandular structures indicating a tubule formation score of 1 in this breast carcinoma
      \item the infiltrating carcinoma cells display preserved tubular architecture resembling normal breast tissue consistent with a tubule formation score of 1
      \item a tubule formation score of 1 is evident on this HE‐stained image showcasing numerous well-formed tubules within the breast carcinoma
      \item this region of invasive breast carcinoma depicted on an HE histopathology image exhibits a high degree of tubular differentiation (tubule formation score 1)
    \end{itemize} \\

  & 2 &
    \begin{itemize}[noitemsep,topsep=0pt,leftmargin=*]
      \item this HE histopathology image of invasive breast carcinoma shows a moderate degree of tubule formation (score 2)
      \item while some tubular structures are present this microscopic view highlights a less organized pattern of growth within the breast carcinoma aligning with a tubule formation score of 2
      \item a mix of both well-formed tubules and areas lacking clear glandular structures suggests a tubule formation score of 2 in this breast carcinoma
      \item the presence of some discernible tubules alongside regions of less-defined glandular architecture indicates a tubule formation score of 2
      \item this invasive breast carcinoma displays an intermediate level of tubular differentiation consistent with a tubule formation score of 2
    \end{itemize} \\

  & 3 &
    \begin{itemize}[noitemsep,topsep=0pt,leftmargin=*]
      \item tubular structures are largely absent within this region of invasive breast carcinoma indicating a tubule formation score of 3 on an HE histopathology image
      \item this breast carcinoma demonstrates disorganized growth with minimal gland formation consistent with a tubule formation score of 3
      \item poorly differentiated tumor cells lacking discernible tubules predominate in this microscopic view suggestive of a tubule formation score of 3
      \item a tubule formation score of 3 is evident within this HE‐stained image where the invasive carcinoma shows a scarcity of well-defined glandular structures
      \item limited tubule formation characterizes this region of breast carcinoma resulting in a tubule formation score of 3
    \end{itemize} \\

  \addlinespace
  \multicolumn{3}{@{}l}{\textbf{Cervical Dysplasia}} \\

  & 1 &
    \begin{itemize}[noitemsep,topsep=0pt,leftmargin=*]
      \item HE‐stained image of a cervical biopsy demonstrating normal squamous epithelium and underlying stroma
      \item microscopic view of a cervical biopsy showing regular stratification of the squamous epithelium without atypia
      \item region of a cervical biopsy with well-defined mature squamous epithelium and unremarkable stroma
      \item cervical biopsy demonstrating normal endocervical glands and adjacent squamous epithelium
      \item HE histopathology of a cervical biopsy showcasing a section of the transformation zone with no pathologic findings
    \end{itemize} \\

  & 2 &
    \begin{itemize}[noitemsep,topsep=0pt,leftmargin=*]
      \item HE‐stained image of a cervical biopsy demonstrating features of cervical intraepithelial neoplasia grade 1 (CIN 1) including nuclear atypia in the lower third of the epithelium
      \item microscopic view of a cervical biopsy with CIN 1 showing enlarged nuclei and increased nuclear-to-cytoplasmic ratio in the basal layer of the epithelium
      \item region of a cervical biopsy exhibiting CIN 1 with mild nuclear atypia and koilocytic change (perinuclear halos)
      \item cervical biopsy demonstrating CIN 1 with subtle nuclear abnormalities confined to the lower epithelial layers
      \item HE histopathology of a cervical biopsy with a focus on CIN 1 characterized by slight disorganization of the squamous epithelium
    \end{itemize} \\

  & 3 &
    \begin{itemize}[noitemsep,topsep=0pt,leftmargin=*]
      \item HE‐stained image of a cervical biopsy demonstrating high-grade dysplasia with significant nuclear atypia and architectural disarray
      \item microscopic view of a cervical biopsy showcasing high-grade squamous intraepithelial lesion (HSIL) with loss of epithelial maturation and cellular disorganization
      \item region of a cervical biopsy exhibiting high-grade dysplasia characterized by increased mitotic activity, nuclear pleomorphism, and loss of normal epithelial polarity
      \item cervical biopsy demonstrating high-grade dysplasia with marked cellular atypia
      \item HE histopathology of a cervical biopsy focusing on high-grade dysplasia showing prominent nuclear abnormalities and disruption of the normal epithelial architecture
    \end{itemize} \\

  \addlinespace
  \multicolumn{3}{@{}l}{\textbf{Prostate Cancer Needle Core Biopsy}} \\

  & 1 &
    \begin{itemize}[noitemsep,topsep=0pt,leftmargin=*]
      \item HE histopathology image showing benign prostate tissue within a prostate specimen
      \item prostate image demonstrating normal prostatic glands and stroma consistent with benign tissue
      \item microscopic view of prostate with well-defined regularly spaced glands and intervening stroma indicative of benign prostatic tissue
      \item focus on a region of benign prostate tissue within a prostate specimen exhibiting a normal glandular architecture and stromal component
      \item HE section of prostate showcasing benign prostatic glands lined by a bilayer epithelium (basal and luminal cells)
    \end{itemize} \\

  & 2 &
    \begin{itemize}[noitemsep,topsep=0pt,leftmargin=*]
      \item HE histopathology image demonstrating Gleason pattern 3 prostate cancer within a prostate specimen
      \item prostate image highlighting individual well-formed glands characteristic of Gleason pattern 3 carcinoma
      \item microscopic view of a prostate with discrete uniformly sized and shaped glands consistent with Gleason pattern 3 prostate cancer
      \item focus on Gleason pattern 3 prostate cancer within prostate specimen showing relatively distinct glandular structures with minimal variability
      \item HE section of prostate demonstrating areas of Gleason pattern 3 adenocarcinoma marked by separate glands composed of tumor cells infiltrating between benign glands
    \end{itemize} \\

  & 3 &
    \begin{itemize}[noitemsep,topsep=0pt,leftmargin=*]
      \item HE histopathology image demonstrating Gleason pattern 4 prostate cancer within a prostate specimen
      \item prostate image highlighting areas of fused glands or poorly formed glands consistent with Gleason pattern 4 carcinoma
      \item microscopic view of prostate showcasing irregular glandular structures with varying shapes and sizes indicative of Gleason pattern 4 prostate cancer
      \item focus on Gleason pattern 4 prostate cancer within a prostate specimen demonstrating loss of normal glandular architecture and a more disorganized growth pattern
      \item HE section of prostate demonstrating regions of Gleason pattern 4 adenocarcinoma characterized by fused or poorly formed glands
    \end{itemize} \\

  & 4 &
    \begin{itemize}[noitemsep,topsep=0pt,leftmargin=*]
      \item HE histopathology image demonstrating Gleason pattern 5 prostate cancer within a prostate specimen
      \item prostate image highlighting sheets of tumor cells lacking any glandular formation characteristic of Gleason pattern 5 carcinoma
      \item microscopic view of prostate showing solid areas of tumor cells representing Gleason pattern 5 prostate cancer
      \item focus on Gleason pattern 5 prostate cancer within a prostate specimen demonstrating poorly differentiated cells and absence of gland formation
      \item HE section of prostate with regions of Gleason pattern 5 adenocarcinoma characterized by a complete lack of glandular differentiation and presence of solid tumor cell clusters
    \end{itemize} \\

  \addlinespace
  \multicolumn{3}{@{}l}{\textbf{Radical Prostatectomy}} \\

  & 1 &
    \begin{itemize}[noitemsep,topsep=0pt,leftmargin=*]
      \item HE histopathology image showing benign prostate tissue
      \item small region from prostate specimen demonstrating normal prostatic glands and stroma consistent with benign tissue
      \item microscopic view of a prostate specimen with well-defined regularly spaced glands and intervening stroma indicative of benign prostatic tissue
      \item focus on a region of benign prostate tissue exhibiting a normal glandular architecture and stromal component
      \item HE section showcasing benign prostatic glands lined by a bilayer epithelium (basal and luminal cells)
    \end{itemize} \\

  & 2 &
    \begin{itemize}[noitemsep,topsep=0pt,leftmargin=*]
      \item HE histopathology image demonstrating Gleason pattern 3 prostate cancer
      \item small region of prostate on histopathology highlighting individual well-formed glands characteristic of Gleason pattern 3 carcinoma
      \item microscopic view of a prostate tissue with discrete uniformly sized and shaped glands consistent with Gleason pattern 3 prostate cancer
      \item focus on Gleason pattern 3 prostate cancer showing relatively distinct glandular structures with minimal variability
      \item HE section of prostate tissue demonstrating areas of Gleason pattern 3 adenocarcinoma marked by separate infiltrating glands
    \end{itemize} \\

  & 3 &
    \begin{itemize}[noitemsep,topsep=0pt,leftmargin=*]
      \item HE histopathology image demonstrating Gleason pattern 4 prostate cancer
      \item small region of a prostate on histopathology highlighting areas of fused glands, poorly formed glands, or cribriform architecture consistent with Gleason pattern 4 carcinoma
      \item microscopic view of prostate showcasing irregular glandular structures with varying shapes and sizes indicative of Gleason pattern 4 prostate cancer
      \item focus on Gleason pattern 4 prostate cancer demonstrating loss of normal glandular architecture and a more disorganized growth pattern
      \item HE section of prostate tissue demonstrating regions of Gleason pattern 4 adenocarcinoma characterized by fused or poorly defined glands
    \end{itemize} \\

  & 4 &
    \begin{itemize}[noitemsep,topsep=0pt,leftmargin=*]
      \item HE histopathology image demonstrating Gleason pattern 5 prostate cancer
      \item small region of prostate on histopathology highlighting presence of sheets of tumor cells lacking any glandular formation characteristic of Gleason pattern 5 carcinoma
      \item microscopic view of prostate showing solid areas of tumor cells with minimal cellular pleomorphism representing Gleason pattern 5 prostate cancer
      \item focus on Gleason pattern 5 prostate cancer demonstrating poorly differentiated cells and absence of gland formation
      \item HE section of prostate with regions of Gleason pattern 5 adenocarcinoma characterized by a complete lack of glandular differentiation and presence of solid tumor cell clusters
    \end{itemize} \\

  \addlinespace
  \multicolumn{3}{@{}l}{\textbf{The Cancer Genome Atlas (TCGA) Study Types}} \\

  & 1 &
    \begin{itemize}[noitemsep,topsep=0pt,leftmargin=*]
      \item microscopic view of bladder specimen demonstrating bladder cancer
      \item microscopic view of bladder specimen demonstrating urothelial cell carcinoma
      \item microscopic view of bladder specimen demonstrating urothelial cancer
      \item microscopic view of bladder specimen demonstrating urinary bladder tumor
      \item microscopic view of bladder specimen demonstrating transitional cell carcinoma
    \end{itemize} \\

  & 2 &
    \begin{itemize}[noitemsep,topsep=0pt,leftmargin=*]
      \item microscopic view of breast specimen demonstrating breast cancer
      \item microscopic view of breast specimen demonstrating invasive breast carcinoma
      \item microscopic view of breast specimen demonstrating carcinoma of the breast
    \end{itemize} \\

  & 3 &
    \begin{itemize}[noitemsep,topsep=0pt,leftmargin=*]
      \item microscopic view of colon specimen demonstrating colon cancer
      \item microscopic view of colon specimen demonstrating colorectal cancer
      \item microscopic view of colon specimen demonstrating colon adenocarcinoma
    \end{itemize} \\

  & 4 &
    \begin{itemize}[noitemsep,topsep=0pt,leftmargin=*]
      \item microscopic view of head and neck squamous cell carcinoma
    \end{itemize} \\

  & 5 &
    \begin{itemize}[noitemsep,topsep=0pt,leftmargin=*]
      \item microscopic view of kidney specimen demonstrating kidney renal clear cell carcinoma
      \item microscopic view of kidney specimen demonstrating clear cell carcinoma
      \item microscopic view of kidney specimen demonstrating renal clear cell carcinoma
      \item microscopic view of kidney specimen demonstrating renal cell carcinoma clear cell type
      \item microscopic view of kidney specimen demonstrating RCC
      \item microscopic view of kidney specimen demonstrating kidney cancer
    \end{itemize} \\

  & 6 &
    \begin{itemize}[noitemsep,topsep=0pt,leftmargin=*]
      \item microscopic view of liver specimen demonstrating liver hepatocellular carcinoma
      \item microscopic view of liver specimen demonstrating liver cancer
      \item microscopic view of liver specimen demonstrating liver carcinoma
      \item microscopic view of liver specimen demonstrating hepatocellular carcinoma
    \end{itemize} \\

  & 7 &
    \begin{itemize}[noitemsep,topsep=0pt,leftmargin=*]
      \item microscopic view of lung specimen demonstrating lung adenocarcinoma
      \item microscopic view of lung specimen demonstrating NSCLC adenocarcinoma
      \item microscopic view of lung specimen demonstrating lung adenocarcinoma
    \end{itemize} \\

  & 8 &
    \begin{itemize}[noitemsep,topsep=0pt,leftmargin=*]
      \item microscopic view of lung specimen demonstrating lung squamous cell carcinoma
      \item microscopic view of lung specimen demonstrating NSCLC squamous carcinoma
      \item microscopic view of lung specimen demonstrating squamous cell lung cancer
    \end{itemize} \\

  & 9 &
    \begin{itemize}[noitemsep,topsep=0pt,leftmargin=*]
      \item microscopic view of ovary specimen demonstrating ovarian cancer
      \item microscopic view of ovary specimen demonstrating ovarian serous cancer
      \item microscopic view of ovary specimen demonstrating serous carcinoma
      \item microscopic view of ovary specimen demonstrating ovarian serous cystadenocarcinoma
    \end{itemize} \\

  & 10 &
    \begin{itemize}[noitemsep,topsep=0pt,leftmargin=*]
      \item microscopic view of stomach specimen demonstrating stomach adenocarcinoma
      \item microscopic view of stomach specimen demonstrating stomach cancer
      \item microscopic view of stomach specimen demonstrating gastric cancer
      \item microscopic view of stomach specimen demonstrating gastric carcinoma
    \end{itemize} \\

  \addlinespace
  \multicolumn{3}{@{}l}{\textbf{Tissue Types}} \\

  & 1 &
    \begin{itemize}[noitemsep,topsep=0pt,leftmargin=*]
      \item HE histopathology image from the appendix
      \item microscopic region of a cross-section of the appendix
      \item microscopic view of the appendix
      \item portion of the appendix on histopathology
      \item HE‐stained image of the appendix showing the tubular structure
    \end{itemize} \\

  & 2 &
    \begin{itemize}[noitemsep,topsep=0pt,leftmargin=*]
      \item HE histopathology image of breast tissue
      \item microscopic view of breast tissue
      \item region of breast tissue on histopathology
      \item HE section of breast tissue
    \end{itemize} \\

  & 3 &
    \begin{itemize}[noitemsep,topsep=0pt,leftmargin=*]
      \item HE histopathology image of the cervix
      \item microscopic view of the cervix demonstrating epithelial cells
      \item cervical tissue on histopathology
      \item region of HE‐stained section of the cervix
    \end{itemize} \\

  & 4 &
    \begin{itemize}[noitemsep,topsep=0pt,leftmargin=*]
      \item HE histopathology image of the colon
      \item colonic mucosa region on histopathology
      \item microscopic view of the colon
      \item HE‐stained image highlighting portion of colon tissue
      \item colon tissue on histopathology
    \end{itemize} \\

  & 5 &
    \begin{itemize}[noitemsep,topsep=0pt,leftmargin=*]
      \item HE histopathology image of the fallopian tube
      \item microscopic cross-section of the fallopian tube
      \item microscopic view of the fallopian tube
      \item HE‐stained section of the fallopian tube
      \item fallopian tube tissue on histopathology
    \end{itemize} \\

  & 6 &
    \begin{itemize}[noitemsep,topsep=0pt,leftmargin=*]
      \item HE histopathology image of the gallbladder
      \item microscopic view of the gallbladder
      \item HE‐stained section showing a region of the gallbladder
      \item gallbladder tissue on histopathology
    \end{itemize} \\

  & 7 &
    \begin{itemize}[noitemsep,topsep=0pt,leftmargin=*]
      \item HE histopathology image of the liver
      \item microscopic view of liver tissue demonstrating hepatocytes arranged in cords
      \item liver tissue on histopathology
    \end{itemize} \\

  & 8 &
    \begin{itemize}[noitemsep,topsep=0pt,leftmargin=*]
      \item HE histopathology image of a lymph node
      \item microscopic view of a lymph node
      \item lymph node demonstrating mixture of immune cells predominantly lymphocytes
      \item HE‐stained section of a lymph node
      \item lymph node tissue on histopathology
    \end{itemize} \\

  & 9 &
    \begin{itemize}[noitemsep,topsep=0pt,leftmargin=*]
      \item HE histopathology image of the ovary
      \item microscopic view of the ovary
      \item microscopic region of ovarian tissue
      \item HE‐stained section of the ovary
      \item ovarian tissue on histopathology
    \end{itemize} \\

  & 10 &
    \begin{itemize}[noitemsep,topsep=0pt,leftmargin=*]
      \item HE histopathology image of the placenta
      \item placental tissue showcasing chorionic villi
      \item microscopic view of the placenta
      \item HE‐stained section of the placenta
      \item placental tissue on histopathology
    \end{itemize} \\

  & 11 &
    \begin{itemize}[noitemsep,topsep=0pt,leftmargin=*]
      \item HE histopathology image of the prostate
      \item microscopic view of prostate tissue
      \item small region of prostate tissue with acini and stroma
      \item HE‐stained section of the prostate
      \item prostate tissue on histopathology
    \end{itemize} \\

  & 12 &
    \begin{itemize}[noitemsep,topsep=0pt,leftmargin=*]
      \item HE histopathology image of the skin
      \item microscopic view of the skin
      \item HE‐stained section of the skin
      \item skin tissue on histopathology
    \end{itemize} \\

  & 13 &
    \begin{itemize}[noitemsep,topsep=0pt,leftmargin=*]
      \item HE histopathology image of the thyroid
      \item microscopic view of the thyroid
      \item HE‐stained section of the thyroid
      \item thyroid tissue on histopathology
    \end{itemize} \\

  & 14 &
    \begin{itemize}[noitemsep,topsep=0pt,leftmargin=*]
      \item HE histopathology image of the upper GI tract such as the stomach
      \item upper GI mucosa region on histopathology
      \item microscopic view of the upper GI tract
      \item HE‐stained image highlighting portion of upper GI tract tissue
      \item upper GI tract tissue on histopathology
      \item stomach or small intestine tissue on histopathology
    \end{itemize} \\

  & 15 &
    \begin{itemize}[noitemsep,topsep=0pt,leftmargin=*]
      \item HE histopathology image of the uterus
      \item microscopic view of the cells of the uterus
      \item HE‐stained section of the uterus
      \item uterine tissue on histopathology
    \end{itemize} \\

  & 16 &
    \begin{itemize}[noitemsep,topsep=0pt,leftmargin=*]
      \item HE histopathology image of the vas deferens
      \item microscopic view of the vas deferens
      \item HE‐stained section of the vas deferens
      \item vas deferens tissue on histopathology
    \end{itemize} \\

\end{longtable}

\begin{table}[htb]
\centering
\footnotesize
\caption{Zero shot prompts for text-based medical evaluations.}
\label{tab:text_eval_prompts}
\begin{tabularx}{\textwidth}{@{}lX@{}}
\toprule
\textbf{Dataset} & \textbf{Prompt} \\
\midrule
MedQA & "Instructions: The following are multiple choice questions about medical knowledge. Solve them in a step-by-step fashion, starting by summarizing the available information. Output a single option from the four options as the final answer. Question: " + \texttt{<QUESTION>} + " Response (think step by step and then end with "Final Answer:" followed by *only* the letter corresponding to the correct answer enclosed in parentheses)"\\
\midrule
PubMedQA & "Instructions: The following are multiple choice questions about medical knowledge. Solve them in a step-by-step fashion, starting by summarizing the available information. Output a single option from the four options as the final answer. Answer the following question given the context (reply with one of the options): Context: " + \texttt{<CONTEXT>} " + Question: " + \texttt{<QUESTION>} + " Response (think step by step and then end with "Final Answer:" followed by *only* the letter corresponding to the correct answer enclosed in parentheses)"\\
\midrule
MedMCQA & "Instructions: The following are multiple choice questions about medical knowledge. Solve them in a step-by-step fashion, starting by summarizing the available information. Output a single option from the four options as the final answer. Question: " + \texttt{<QUESTION>} + "Response (think step by step and then end with "Final Answer:" followed by *only* the letter corresponding to the correct answer enclosed in parentheses)"\\
\midrule
MMLU & "Instructions: The following are multiple choice questions about medical knowledge. Solve them in a step-by-step fashion, starting by summarizing the available information. Output a single option from the four options as the final answer. Question: " + \texttt{<QUESTION>} + "  Response (think step by step and then end with "Final Answer:" followed by *only* the letter corresponding to the correct answer enclosed in parentheses)"\\
\midrule
MedXpertQA & \texttt{<QUESTION>} + " Response (think step by step through each of the multiple choice options. You MUST end your response with `Final Answer:' followed by *only* the letter corresponding to the correct answer enclosed in parentheses) like `Final Answer:(X)'." \\
\bottomrule
\end{tabularx}
\end{table}

\begin{table}[htb]
\centering
\footnotesize
\caption{Zero-shot prompts for image classification and VQA evaluations. For all but the medical MedGemma and Med-Gemini models, the prompts below were prefixed with the following system message (\texttt{SM}): \texttt{MI}:~"You are a helpful medical assistant.", or \texttt{RI}:~"You are a helpful radiology assistant.", as indicated.}
\label{tab:classification_eval_prompts}
\begin{tabularx}{\textwidth}{@{}p{0.1\textwidth}lX@{}}
\toprule
\textbf{Dataset(s)} & \textbf{SM} & \textbf{Prompt/Example} \\
\midrule

MIMIC-CXR, CheXpert & \texttt{MI} & \texttt{<IMAGE>} + " Is there \texttt{<CONDITION>} in this image? You may write out your argument before stating your final very short, definitive, and concise answer (if possible, a single word or the letter corresponding to your answer choice) X in the format "Final Answer: X":"{\vspace{\bigskipamount}} 

\vspace{\medskipamount}
Where \texttt{<CONDITION>} is one of the following:
\begin{itemize}[nosep, leftmargin=1.5em, itemsep=2pt]
    \item \textbf{For MIMIC-CXR \& CheXpert}: \texttt{atelectasis}, \texttt{cardiomegaly}, \texttt{consolidation}, \texttt{edema}, \texttt{pleural effusion}.
    \item \textbf{For CXR14}: \texttt{lung opacity}, \texttt{pneumothorax}.
    \item[] \textit{Exception for CXR14}: The \texttt{fracture} condition uses the question "Are there any fractures in the image?".
\end{itemize}\\
\midrule
PathMCQA & MI & "\texttt{<QUESTION>}"\vspace{\baselineskip}

Example for \texttt{<QUESTION>}:

"What is the predominant diagnostic finding in this cervical biopsy image? Please select the correct option from the following choices: A) Normal or benign tissue B) Mild dysplasia (CIN1) C) Moderate or severe dysplasia (CIN 2+)" \\
\midrule
US-Derm MCQA & \texttt{MI} & \texttt{<IMAGE>} + "\texttt{<QUESTION>} You may write out your argument before deciding on the most likely condition X (with X being one of: A, B, C, or D) in the format: "The most likely diagnosis is: X":" \vspace{\bigskipamount}

Example for \texttt{<QUESTION>}:

"Question: Given the following image, what is the most likely dermatology condition? Options: (A) Granuloma annulare. (B) Melanocytic Nevus. (C) Erythema annulare centrifugum. (D) Morphea/Scleroderma."\\
\midrule

EyePACS & \texttt{MI} &\texttt{<IMAGE>} + "Given this fundus image, determine the most likely diabetic retinopathy (DR) stage present, even if you are unsure:

\hspace*{1em}A: No DR
  
\hspace*{1em}B: mild DR
  
\hspace*{1em}C: moderate DR
  
\hspace*{1em}D: severe DR
  
\hspace*{1em}E: proliferative DR
  
You may write out your argument before stating your final, very short, definitive, and concise answer (no more than a few words) and letter corresponding to your answer choice X in the format "The most likely diagnosis is: X":"\\
\midrule

SLAKE & \texttt{MI} & \texttt{<IMAGE>} + \texttt{<QUESTION>} + "You may write out your argument before stating your final very short, definitive, and concise answer (if possible, a single word or the letter corresponding to your answer choice) X in the format "Final Answer: X": " \\
\midrule

VQA-Rad & \texttt{RI} & \texttt{<IMAGE>} + "Given this radiology image, which can be a frontal chest X-ray, a single slice head or abdominal CT or MR image, provide a very short, definitive, and concise answer (if possible, a single word) to the following question: \texttt{<QUESTION>}" \\
\midrule

MedXpertQA (multi-modal questions) & \texttt{MI} & \texttt{"Figure A" + <IMAGE\_A> + "Figure B" + <IMAGE\_B> + ... +} "\texttt{<QUESTION>} Think step by step through each of the multiple choice options. You MUST end your response with 'Final Answer:' followed by only the letter corresponding to the correct answer enclosed in parentheses)."\\
\midrule

CXR report generation & \texttt{-} & \texttt{<IMAGE>} "\texttt{<INDICATION>} findings:"\vspace{\baselineskip}

Where \texttt{<INDICATION>} is the indication section of the report as prefix.\\
\bottomrule
\end{tabularx}
\end{table}

\clearpage

\section{Comparison of CXR data-efficient learning}

\begin{figure}[htp]
    \centering
    \includegraphics[width=0.9\linewidth]{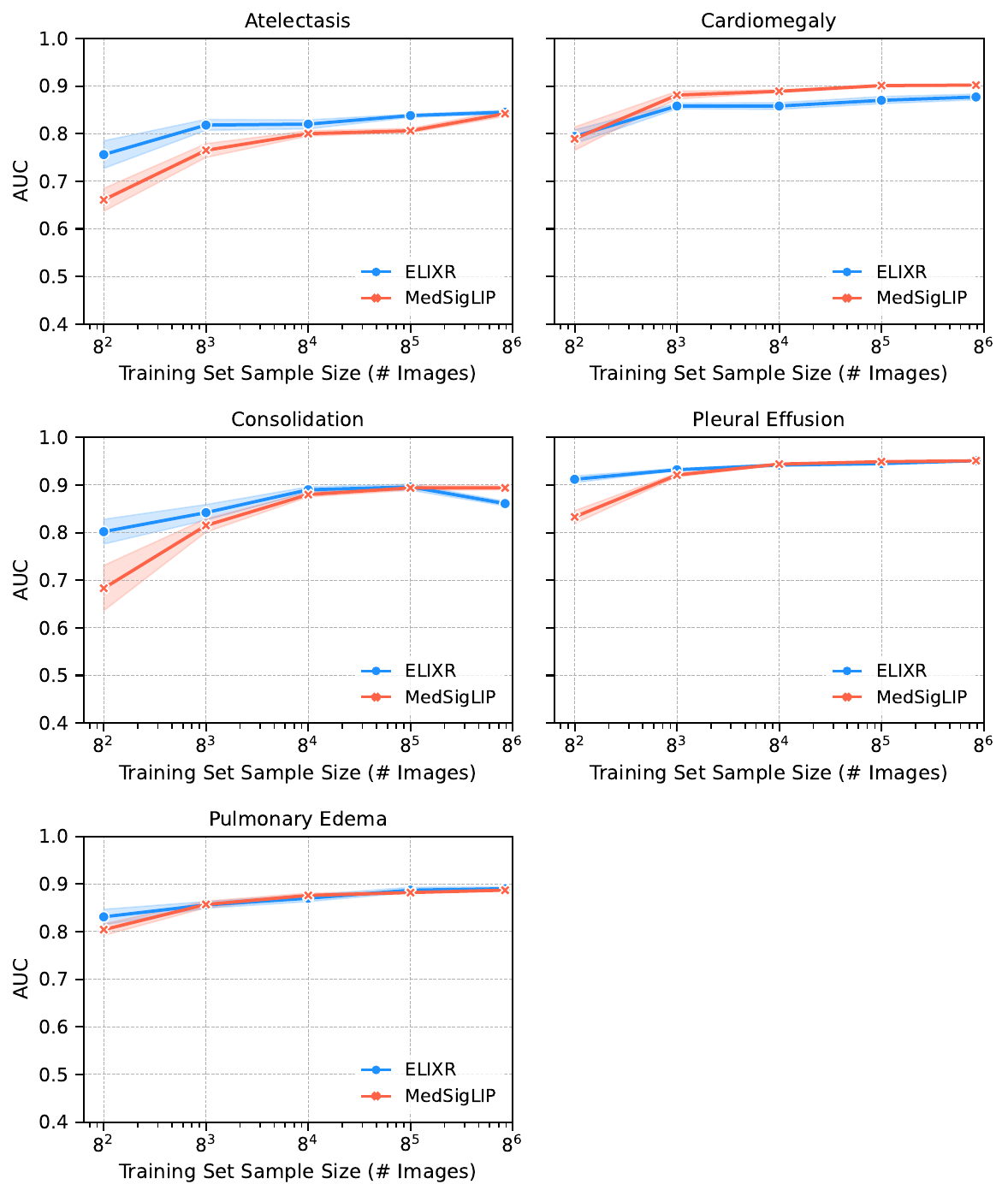}
    \caption{\small{\textbf{Individual results for CXR data efficient learning on CheXpert datasets comparing to ELIXR~\cite{xu2023elixr} 
    }}}
    \label{fig:cxr_chexpert_metric}
\end{figure}

\begin{figure}[htp]
    \centering
    \includegraphics[width=0.9\linewidth]{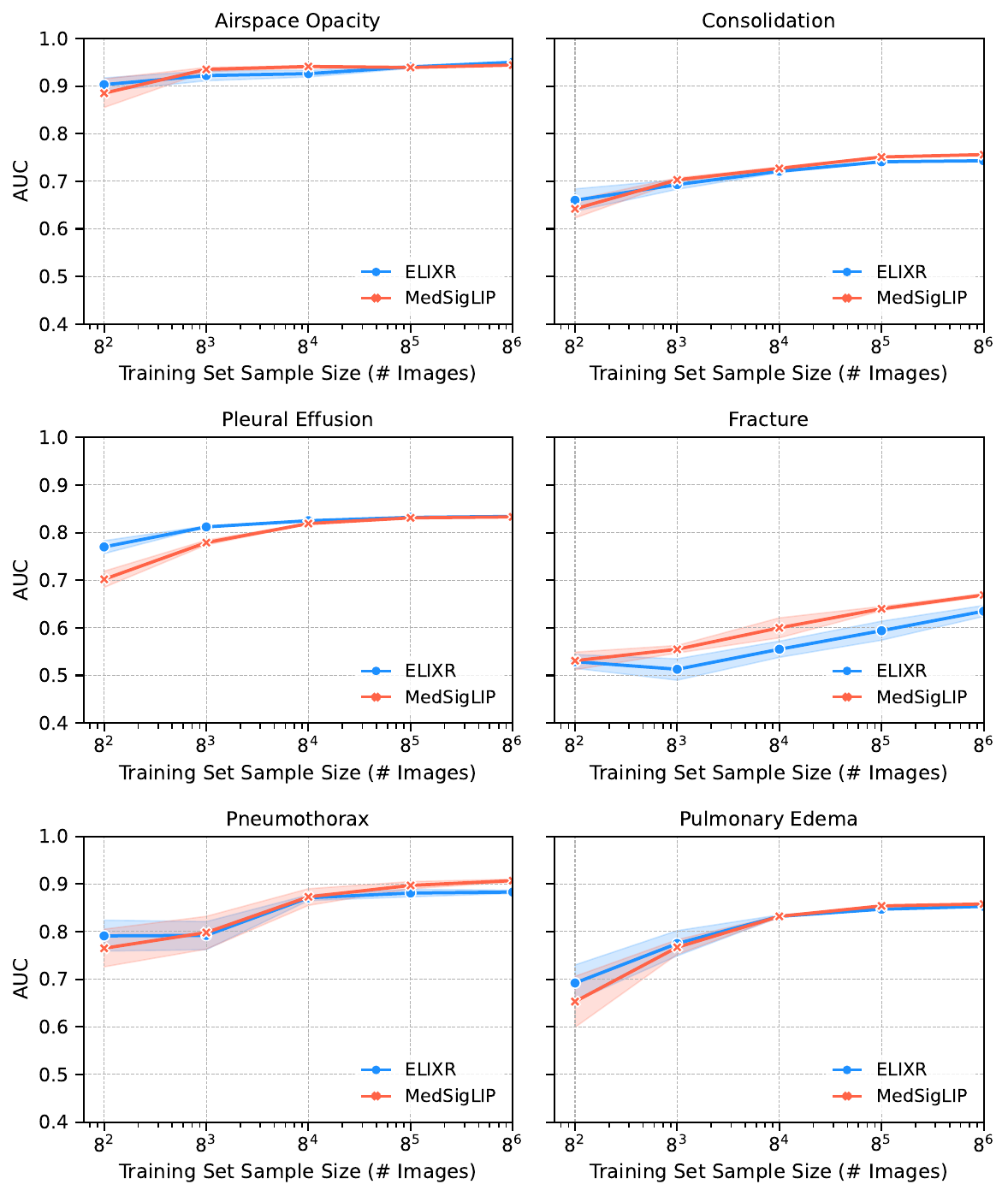}
    \caption{\small{\textbf{Individual results for CXR data efficient learning on CXR14 datasets comparing to ELIXR~\cite{xu2023elixr} 
    }}}
    \label{fig:cxr_cxr14_metric}
\end{figure}

To compare with previous research on CXR image encoder-based data-efficient learning, we compared MedSigLIP to ELIXR \cite{xu2023elixr}. Performance of MedSigLIP and ELIXR are compared for various CXR findings at different training set sizes on CheXpert (Figure~\ref{fig:cxr_chexpert_metric}) and CXR14 (Figure~\ref{fig:cxr_cxr14_metric}).

\clearpage

\section{Additional details about EHRQA}

\begin{figure}[!hp]
    \centering
        \includegraphics[width=0.75\linewidth]{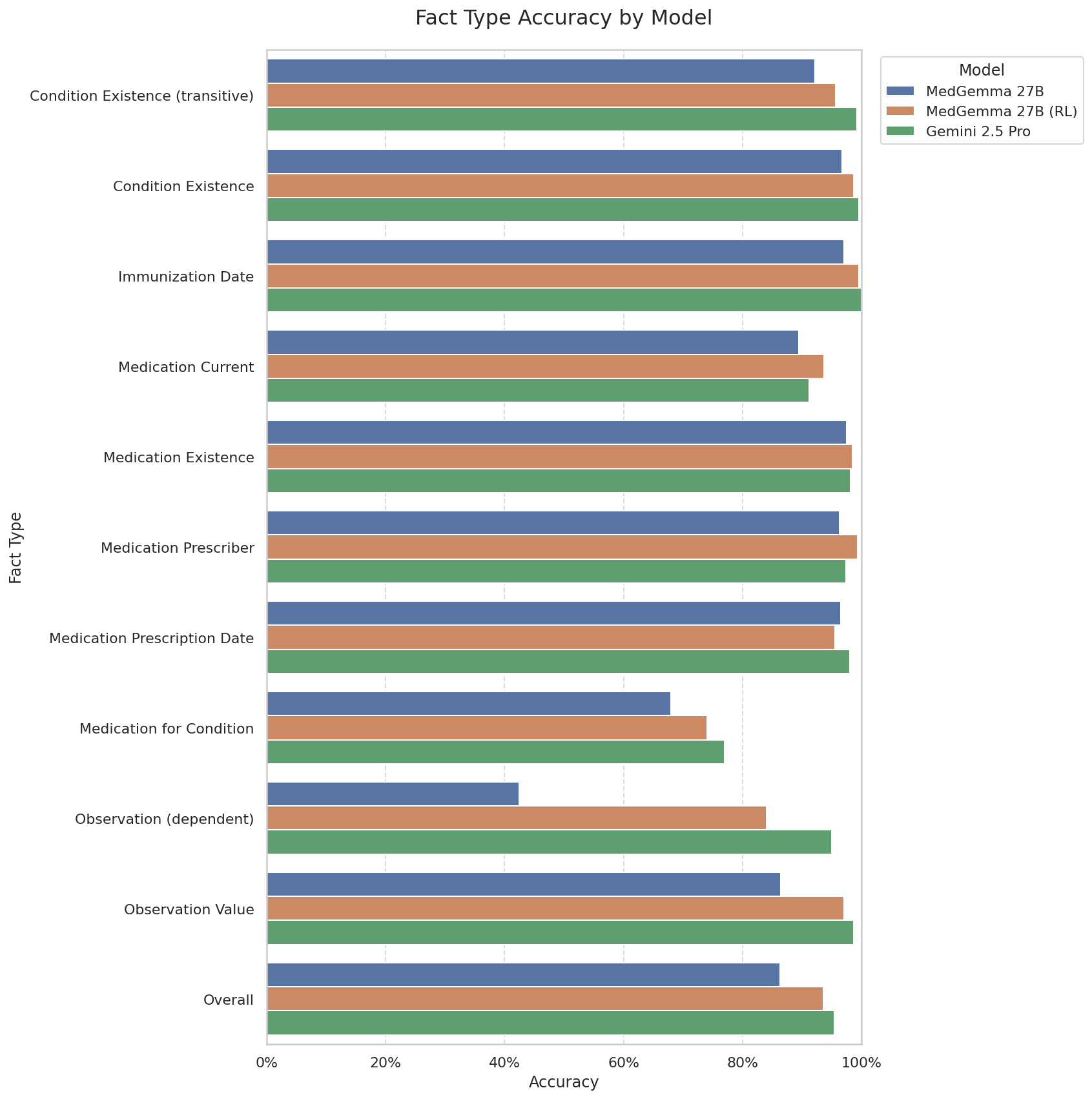}
    
    \caption{EHRQA detailed results, comparing MedGemma 27B (text-only) before and after RL-tuning, along with Gemini 2.5 Pro}
    \label{fig:ehrqa_results_figure}
\end{figure}

EHRQA detailed results appear in Figure~\ref{fig:ehrqa_results_figure}, where the base MedGemma 27B text model is compared against the fine-tuned MedGemma 27B model and Gemini 2.5 Pro. Details of EHRQA question categories are shown in Table~\ref{tab:ehrqa_examples} and prompts for question rewrites and evaluation are shown in Table~\ref{tab:ehrqa_prompts}.

\begin{table*}[htb]
\centering
\caption{EHRQA Question Categories with Original (Template-Generated) and LLM-Rephrased Examples.}
\label{tab:ehrqa_examples}
\small
\begin{tabular}{
    >{\RaggedRight}p{0.16\textwidth}
    >{\RaggedRight}p{0.24\textwidth}
    >{\RaggedRight}p{0.24\textwidth}
    >{\RaggedRight\arraybackslash}p{0.24\textwidth} 
}
\toprule
\textbf{Category} & \textbf{Description} & \textbf{Example: Templated} & \textbf{Example: LLM rephrased} \\
\midrule

Condition existence (transitive) & Questions related to whether a patient has a history of a general class of medical conditions. & Has the patient had a medical condition in gingival structure in the past? & Has the patient had a past medical condition affecting the gums? \\
\midrule

Condition existence & Questions related to whether a patient has a history of a given medical condition. & Has the patient had acute bronchitis in the past? & Has the patient had acute bronchitis previously? \\
\midrule

Immunization date & Questions related to the existence of patient immunizations and their immunization dates. & When did the patient last receive Hep B, adult immunization? & When was the patient's last adult hepatitis B vaccination administered? \\
\midrule

Medication current & Questions related to whether the patient has active prescriptions for a given medication. & Does the patient have a prescription for \{28 (norethindrone 0.35 MG Oral Tablet) \} Pack? & Does the patient have an active prescription for norethindrone oral tablet? \\
\midrule

Medication existence & Questions related to whether the patient has ever taken a given medication or type of medication. & Did the patient have a prescription for insulin isophane, human 70 UNT/ML / insulin, regular, human 30 UNT/ML Injectable Suspension in the past? & Did the patient have a prescription for human insulin isophane / human insulin regular injectable suspension in the past? \\
\midrule

Medication prescriber & Questions related to the prescriber of a given medication for a patient. & Who prescribed acetaminophen 325 MG Oral Tablet on the date 2022-06-30? & Who prescribed acetaminophen on June 30, 2022? \\
\midrule

Medication prescription date & Questions related to the medication prescription dates. & When was acetaminophen 325 MG Oral Tablet first prescribed? & When was acetaminophen oral tablet first prescribed? \\
\midrule

Medication for condition & Questions on the relationships between medication and medical condition history. & Which condition were the medications acetaminophen 300 MG / hydrocodone bitartrate 5 MG Oral Tablet prescribed for? & For what condition was acetaminophen/hydrocodone prescribed? \\
\midrule

Observation (dependent) & Questions related to patient lab observations related to some medical event. & What was the patient's first recorded value for Pain severity - 0-10 verbal numeric rating [Score] - Reported after they were first prescribed Acetaminophen 325 MG Oral Tablet? & What was the patient's initial reported pain score after the first prescription of acetaminophen 325 mg oral tablet? \\
\midrule

Observation value & Questions related to patient lab observations and their values. & What is the patient's most recent recorded value for Heart rate? & What is the patient's most recent heart rate reading? \\
\bottomrule
\end{tabular}
\end{table*}

\begin{table}[!hp]
\centering
\footnotesize
\caption{Prompts for EHRQA construction and evaluation.}
\label{tab:ehrqa_prompts}
\begin{tabularx}{\textwidth}{@{}lX@{}}
\toprule
\textbf{Purpose} & \textbf{Prompt} \\
\midrule
\textbf{General question rewrite} & {
You are a helpful medical editing assistant.

You will be presented with a question which pertains to the medical record for
a patient. Rewrite the question in a way that sounds more natural, omitting
unnecessary (in particular parenthesized) details but retaining enough detail
to leave the question unambiguous.

- Use correct capitalization and language appropriate for a medical
  professional.

- Avoid contractions in the question, but use common-language terminology
  where that is possible without causing ambiguity.

- Keep in mind that the question may be a multiple choice question followed by
  several answer choices.

- Be sure the rewritten questions ask about the same topic as the original. For
  instance, if there is a question about who prescribed a medication,
  the re-written question should do the same.

ORIGINAL QUESTION:
\texttt{<ORIGINAL\_QUESTION>}

REWRITTEN QUESTION:
}\\
\midrule
\textbf{Medication question rewrite} & {
You are a helpful medical editing assistant.

You will be presented with a question which pertains to the medical record for
a patient. Rewrite the question in a way that sounds more natural, omitting
unnecessary (in particular parenthesized) details but retaining enough detail
to leave the question unambiguous.

- Use correct capitalization and language appropriate for a medical
  professional.

- Avoid contractions in the question, but use common-language terminology
  where that is possible without causing ambiguity.

- Keep in mind that the question may be a multiple choice question followed by
  several answer choices.

- If the brand name is mentioned in the original question, rewrite the question using that brand name only.

- Remove the medication strength/dosage information in the medication name.

- Use common names for medication classes if available.

EXAMPLES: \texttt{<FEW\_SHOT\_EXAMPLES>}

ORIGINAL QUESTION:
\texttt{<ORIGINAL\_QUESTION>}

REWRITTEN QUESTION:
}\\
\midrule
\textbf{Evaluation} & {
You are a medical assistant specializing in answer questions about a patient electronic health records (EHR). You will be provided with a patient's full EHR data and a question for you to answer.

======= Patient EHR Data =======

\texttt{<PATIENT\_CONTEXT>}

======= End Patient EHR Data =======

Based on the data provided in the patient's EHR, please answer concisely with no explanation.
If the question is a multiple choice question, answer with only the letter of the correct choice, such as "B". If the question
is a free response question, answer with the correct data. In either case, remember to
not provide an explanation.

Question: \texttt{<QUESTION>}

Answer choices: \texttt{ANSWER\_CHOICES}
}\\
\bottomrule
\end{tabularx}
\end{table}

\clearpage

\section{Additional medical reasoning examples}
\label{sec:additional_med_reasoning_examples}

Tables~\ref{tab:cardiology_text_example_long}, \ref{tab:cardiology_text_example_succinct}, \ref{tab:infectious_text_example_long} and \ref{tab:infectious_text_example_succinct} show additional medical reasoning examples, with clinical commentary, in the same form as Tables~\ref{tab:oncology_text_example_long} and~\ref{tab:oncology_text_example_succinct}.

\renewcommand{\arraystretch}{1.1}
\small

\begin{longtable}{@{} >{\raggedright\arraybackslash}p{\textwidth} @{}}

\caption{Example of MedGemma 27B open clinical reasoning in a cardiology setting, including commentary by a US board-certified practicing cardiologist}
\label{tab:cardiology_text_example_long} \\

\toprule
\textbf{Prompt} \\
\midrule
\endfirsthead

\multicolumn{1}{c}{\tablename\ \thetable\ -- \textit{Continued from previous page}} \\
\toprule

\endhead

\multicolumn{1}{r}{\textit{Continued on next page}} \\
\endfoot

\bottomrule
\endlastfoot

A 55-year old woman with active smoking history, type II diabetes, hypertension, history of coronary artery disease was recently hospitalized for heart failure with an ejection fraction of 35\%. She is presenting for follow-up post discharge. Her discharge medications included lisinopril 10 mg daily, carvedilol 2.5 mg twice daily, and eplerenone 25 mg daily, in addition to her usual medications for diabetes. In clinic, she reports feeling symptomatic with dyspnea after going up 1 flight of stairs though is asymptomatic at rest.

Vital signs at rest: BP 130/90, HR 75, RR 10, Temperature 97.2F, SpO2 98\% on room air. Labs are remarkable for Na 132 mEq/L, K 4.9 mEq/L, BUN 22 mg/dL, Cr 1.04 mg/dL.

The patient reports having a friend on Entresto. Please review whether Entresto is indicated in this patient and why or why not. \\
\midrule
\pagebreak
\textbf{MedGemma 27B output} \\
\midrule
Okay, let's review the indication for Entresto (sacubitril/valsartan) in this patient.

\textbf{Patient Summary:}
\begin{itemize}[nosep, leftmargin=*]
\item \textbf{Diagnosis:} Heart Failure with Reduced Ejection Fraction (HFrEF), EF 35\%.
\item \textbf{Symptoms:} Dyspnea on exertion (1 flight of stairs), NYHA Class II-III.
\item \textbf{Current HFrEF Medications:} Lisinopril (ACE inhibitor), Carvedilol (Beta-blocker), Eplerenone (MRA).
\item \textbf{Comorbidities:} Smoking, Type II Diabetes, Hypertension, CAD.
\item \textbf{Relevant Labs:} K 4.9 mEq/L (high-normal), Cr 1.04 mg/dL (mildly elevated/high-normal).
\item \textbf{Vitals:} BP 130/90 mmHg (acceptable for initiation).
\end{itemize}

\textbf{Entresto Indication Review:}

Entresto (sacubitril/valsartan) is indicated for the reduction of cardiovascular death and hospitalization for heart failure in adult patients with chronic heart failure (NYHA class II-IV) with reduced ejection fraction (LVEF $\leq 40$\%).

\textbf{Analysis for this Patient:}
\begin{enumerate}[nosep, leftmargin=*]
\item \textbf{Ejection Fraction:} The patient's EF is 35\%, which meets the criteria ($\leq 40$\%).
\item \textbf{Symptom Status:} The patient is symptomatic (dyspnea on exertion) despite being on foundational guideline-directed medical therapy (GDMT) including an ACE inhibitor (lisinopril), a beta-blocker (carvedilol), and an MRA (eplerenone). This is the target population for Entresto.
\item \textbf{Contraindications:}
\begin{itemize}[nosep, leftmargin=*]
\item \textbf{ACE Inhibitor Use:} Entresto should \textit{not} be used concurrently with an ACE inhibitor (like lisinopril) due to increased risk of angioedema, hypotension, and hyperkalemia. The lisinopril would need to be stopped, typically with a 36-hour washout period, before starting Entresto.
\item \textbf{Angioedema History:} No history of angioedema related to ACE inhibitors or ARBs is mentioned.
\item \textbf{Hypotension:} Her current BP of 130/90 mmHg is generally acceptable for initiating Entresto, although close monitoring is needed.
\item \textbf{Renal Function:} Her creatinine is 1.04 mg/dL. While Entresto can cause a small increase in creatinine, this level is not a contraindication, but renal function should be monitored.
\item \textbf{Potassium:} Her potassium is 4.9 mEq/L (high-normal). This requires close monitoring, as Entresto can increase potassium levels, especially in combination with an MRA like eplerenone.
\end{itemize}
\end{enumerate}

\textbf{Conclusion:}

\textbf{Yes, Entresto is indicated for this patient.}

\textbf{Reasoning:} She has symptomatic HFrEF (EF 35\%) despite being on appropriate foundational therapy (ACE inhibitor, beta-blocker, MRA). Clinical trials have demonstrated that switching from an ACE inhibitor (like lisinopril) to Entresto significantly reduces morbidity and mortality in this patient population.

\textbf{Important Considerations:}
\begin{itemize}[nosep, leftmargin=*]
\item \textbf{Lisinopril Discontinuation:} The lisinopril must be stopped before starting Entresto.
\item \textbf{Monitoring:} Close monitoring of blood pressure, heart rate, renal function (creatinine/eGFR), and potassium levels is crucial, especially after initiation and dose titration.
\item \textbf{Dosing:} Entresto is typically started at a low dose (e.g., 24/25 mg BID) and titrated upwards to the target dose (e.g., 49/51 mg BID or 97/103 mg BID) as tolerated, based on guideline recommendations.
\end{itemize}

This represents a significant therapeutic opportunity to improve her heart failure outcomes. \\
\midrule
\textbf{Commentary by a US board-certified practicing cardiologist} \\
\midrule
\textbf{Strengths:} Correctly states that entresto is indicated for the reduction of CV death and HF hospitalization. Appropriately checks renal function, electrolytes, angioedema and BP. Lists washout and monitoring.

\textbf{Areas for Improvement:} Dose is incorrect for the lowest dose of entresto. Minor nit that entresto has now been shown in
meta-analysis to reduce all cause mortality in EF<40\% (not just cardiac mortality as stated) \cite{evbayekha2025sacubitril}. Creatinine is mentioned but estimated GFR is the more useful measure.  Minor but no mention of carvedilol dose. The dose is very low and should be increased. If it is increased, there will be less BP room for entresto so would need to titrate both carefully but can be done rapidly (Strong HF trial). The other thing that could have been mentioned is empaglaflozin which is not included in this list. But the question is specific to entresto so understandable. \\

\end{longtable}

\begin{table}[htbp]
\caption{Example of MedGemma 27B open clinical reasoning in a cardiology setting with \emph{succinct} output, including commentary by a US board-certified practicing cardiologist}
\label{tab:cardiology_text_example_succinct}
\renewcommand{\arraystretch}{1.1}

\small

\begin{tabularx}{\textwidth}{@{} >{\raggedright\arraybackslash}X @{}}
\toprule
\textbf{Prompt} \\
\midrule
\emph{Same as in Table~\ref{tab:cardiology_text_example_long} with the suffix: Please be succinct.} \\
\midrule
\textbf{MedGemma 27B output} \\
\midrule
\textbf{Yes, Entresto is indicated.}

\textbf{Rationale:} This patient has symptomatic Heart Failure with reduced Ejection Fraction (HFrEF) (EF 35\%, NYHA Class II) despite being on foundational guideline-directed medical therapy (ACE inhibitor, beta-blocker, MRA). Entresto (sacubitril/valsartan) is recommended to replace the ACE inhibitor (lisinopril) in such patients to further reduce cardiovascular death and heart failure hospitalizations. Her current BP and potassium levels are acceptable for initiation, although close monitoring is required. \\
\midrule
\textbf{Clinical commentary by a US board-certified practicing cardiologist} \\
\midrule
\textbf{Strengths:} A more useful answer - short and to the point, including the rationale for the treatment.

\textbf{Areas for Improvement:} Does not include renal function in the list of things to monitor (just BP and potassium). Same note as above that entresto has now been shown in meta-analysis to reduce all cause mortality in EF<40\% rather than just cardiac mortality as stated. \\
\bottomrule
\end{tabularx}
\end{table}

\begin{longtable}{@{} >{\raggedright\arraybackslash}p{\textwidth} @{}}

\caption{Example of MedGemma 27B open clinical reasoning in an infectious disease setting, including commentary by a UK Specialist Registrar in infectious diseases}
\label{tab:infectious_text_example_long} \\

\toprule
\textbf{Prompt} \\
\midrule
\endfirsthead

\multicolumn{1}{c}{\tablename\ \thetable\ -- \textit{Continued from previous page}} \\
\toprule

\endhead

\multicolumn{1}{r}{\textit{Continued on next page}} \\
\endfoot

\bottomrule
\endlastfoot

A 17-year old otherwise healthy man presents in the emergency room for fatigue and spiking fevers. Two weeks prior, he reports having felt flu-like symptoms with fever, nausea, vomiting, and cough. Symptoms persisted but have worsened, with new onset shortness of breath, headache, and abdominal discomfort. There have been no sick contacts at home. He has been backpacking across various regions in the United States, most recently in Hawaii. 

On evaluation, vital signs are remarkable for SBP 115/85, HR 110, RR 14, Temperature 102.2F, SpO2 90\% on room air. Exam is remarkable for bilateral crackles and icterus.

Labs show the following:

\begin{itemize}[nosep, leftmargin=*]
    \item WBC: 18.2 $\times$ 10$^3$/mcL, Hb: 15.5 g/dL, Plt: 120 $\times$ 10$^3$/mcL
    \item Na: 142 mEq/L, K: 5.1 mEq/L, Cl: 101 mEq/L, CO$_2$: 26 mEq/L, BUN: 46 mg/dL, Cr: 2.2 mg/dL
    \item ALP: 84 U/L, ALT: 175 U/L, AST: 140 U/L, Total bilirubin: 2.0 mg/dL
\end{itemize}

CXR shows bilateral pulmonary infiltrates. A follow-up echocardiogram shows EF 40\%. A respiratory viral panel was negative. Blood, sputum, and urine cultures are taken. 

What would be the next best steps in terms of diagnostic testing to establish the etiology of his symptoms? \\
\midrule
\pagebreak
\textbf{MedGemma 27B output} \\
\midrule
\textbf{Rationale Breakdown:}

\begin{enumerate}[nosep, leftmargin=*]
    \item \textbf{Clinical Presentation Analysis:} The patient presents with a severe, multi-system illness characterized by fever, fatigue, shortness of breath, headache, abdominal discomfort, jaundice (icterus), hypoxia, leukocytosis, thrombocytopenia, acute kidney injury (elevated BUN/Cr), hepatitis (elevated ALT/AST/bilirubin), and possible myocarditis (reduced EF on echo). The illness began with flu-like symptoms two weeks ago and has progressively worsened.

    \item \textbf{Key Clues:}
    \begin{itemize}[nosep, leftmargin=*]
        \item \textbf{Travel History:} Recent travel to Hawaii is a significant clue, pointing towards infections endemic to that region.
        \item \textbf{Multi-organ Involvement:} The combination of pulmonary infiltrates, jaundice, AKI, and myocarditis suggests a systemic process, often infectious.
        \item \textbf{Specific Lab Findings:} Leukocytosis with thrombocytopenia, elevated liver enzymes, and bilirubin are important diagnostic pointers.
    \end{itemize}

    \item \textbf{Differential Diagnosis (Focusing on Hawaii-Endemic Infections):}
    \begin{itemize}[nosep, leftmargin=*]
        \item \textbf{Leptospirosis:} Highly consistent with the presentation. Causes fever, headache, myalgia, conjunctival suffusion (not mentioned but common), jaundice, renal failure, pulmonary hemorrhage/ARDS, and myocarditis. Thrombocytopenia is common; leukocytosis can occur. Hawaii is an endemic area.
        \item \textbf{Dengue Fever:} Endemic in Hawaii. Causes fever, headache, myalgia, rash (not mentioned), leukopenia (patient has leukocytosis), thrombocytopenia. Severe dengue can cause organ impairment, but the specific combination of jaundice, AKI, and myocarditis with leukocytosis makes it less likely than leptospirosis.
        \item \textbf{Hantavirus Pulmonary Syndrome (HPS):} Endemic in Hawaii. Causes fever, myalgia, headache, followed by respiratory failure and pulmonary edema. Thrombocytopenia and leukocytosis are typical. However, significant jaundice and myocarditis are less characteristic of HPS.
        \item \textbf{Rickettsial Infections (e.g., Murine Typhus, Scrub Typhus):} Endemic in Hawaii. Can cause fever, headache, rash, myalgia, sometimes pneumonia, hepatitis, AKI.
        \item \textbf{Other possibilities:} Influenza (severe complications), other viral hepatitis (EBV, CMV), bacterial sepsis, melioidosis (rare but reported).
    \end{itemize}

    \item \textbf{Prioritizing Diagnostic Tests:} Based on the clinical picture and travel history, Leptospirosis is the most likely diagnosis. Therefore, specific testing for Leptospirosis should be prioritized. Dengue and Hantavirus are also important considerations due to endemicity and overlapping features.

    \item \textbf{Specific Diagnostic Tests:}
    \begin{itemize}[nosep, leftmargin=*]
        \item \textbf{Leptospirosis:}
        \begin{itemize}[nosep, leftmargin=*]
            \item \textbf{Serology:} ELISA for IgM antibodies is often the first step for faster results. Microscopic Agglutination Test (MAT) is the gold standard but takes longer. Paired sera (acute and convalescent) are often needed for definitive diagnosis.
            \item \textbf{PCR:} Detects Leptospira DNA in blood (early) or urine (can be positive later, especially after the first week). Given the patient is 2 weeks into symptoms, urine PCR might be useful.
        \end{itemize}
        \item \textbf{Dengue:}
        \begin{itemize}[nosep, leftmargin=*]
            \item \textbf{Serology:} IgM/IgG antibodies.
            \item \textbf{NS1 Antigen:} Detectable early in illness.
            \item \textbf{PCR:} Detects viral RNA early in illness.
        \end{itemize}
        \item \textbf{Hantavirus:}
        \begin{itemize}[nosep, leftmargin=*]
            \item \textbf{Serology:} IgM/IgG antibodies.
            \item \textbf{PCR:} Detects viral RNA early in illness.
        \end{itemize}
        \item \textbf{Other Tests:} While awaiting results for the primary suspects, continue monitoring organ function (liver, kidney), coagulation, and review pending cultures. Consider additional tests like EBV/CMV serology, HIV testing, and Rickettsial serology/PCR if the initial workup is negative or the clinical picture evolves.
    \end{itemize}

    \item \textbf{Conclusion:} The most crucial next steps are to order specific diagnostic tests for the most likely endemic infections given the patient's presentation and travel history, particularly Leptospirosis, Dengue, and Hantavirus.
\end{enumerate}

\textbf{Final Answer:} The final answer is $\boxed{\text{Order specific serological and PCR testing for Leptospirosis, Dengue, and Hantavirus}}$ \\
\midrule
\textbf{Commentary by a UK Specialist Registrar in infectious diseases} \\
\midrule
\textbf{Strengths:} Well-structured answer, making the information easy to parse. Correctly has the most likely diagnosis (lepto) as the focus of the answer. The suggested diagnostic testing enables confirmation of the specific diagnoses listed in the differential.

\textbf{Areas for Improvement:} Dengue fever, hantavirus and melioidosis don’t seem to occur in Hawaii much, if at all (and at least, they are not endemic, as stated in the answer). The suggested diagnostic testing perhaps narrows towards specific diagnoses too quickly, whereas in reality broader testing may still be necessary (e.g. CSF analysis given headache, further chest imaging via CT, etc.) alongside more specific confirmatory tests. It could mention that many of these diagnostic tests are relatively unusual in most hospitals, often requiring samples to be sent via courier to specialized labs, and therefore there is a significant time lag associated with getting results. This may necessitate empiric treatment whilst awaiting for results, which is an important clinical consideration. \\

\end{longtable}

\begin{table}[htbp] 
\caption{Example of MedGemma 27B open clinical reasoning in an infectious disease setting with \emph{succinct} output, including commentary by a UK Specialist Registrar in infectious diseases}
\label{tab:infectious_text_example_succinct}
\renewcommand{\arraystretch}{1.1}

\small

\begin{tabularx}{\textwidth}{@{} >{\raggedright\arraybackslash}X @{}}
\toprule
\textbf{Prompt} \\
\midrule
\emph{Same as in Table~\ref{tab:infectious_text_example_long} with the suffix: Please be succinct.} \\
\midrule
\textbf{MedGemma 27B output} \\
\midrule
\textbf{Next best steps in diagnostic testing:}

\begin{enumerate}[nosep, leftmargin=*]
    \item Leptospirosis serology (IgM ELISA/MAT).
    \item Hantavirus serology (IgM).
    \item Dengue serology (IgM/IgG, NS1 antigen).
    \item Blood and urine cultures.
    \item Urinalysis.
    \item Cardiac enzymes (Troponin).
    \item Coagulation studies.
\end{enumerate}  \\
\midrule
\textbf{Commentary by a UK Specialist Registrar in infectious diseases} \\
\midrule
\textbf{Strengths:} Correctly lists confirmatory diagnostic tests for each suggested differential. Covers both specific confirmatory tests (to pinpoint a particular condition) and more generalized diagnostics (e.g. coagulation studies) that would yield more useful information to guide next steps in case the confirmatory diagnostic tests are negative.

\textbf{Areas for Improvement:} Same feedback as before regarding the epidemiology of hantavirus/dengue in Hawaii - seems to be rare, if at all present. Repeats several items already listed as being performed in the query (e.g. blood, urine cultures). Unclear why it limits the infectious disease testing to a single antibody (IgM) for each potential cause  - it would be normal to perform IgM and IgG as both are useful in distinguishing between an acute infection versus a prior infection. No mention of molecular (genetic) testing which is increasingly used for diagnosis. \\
\bottomrule
\end{tabularx}
\end{table}

\clearpage

\section{MedGemma 27B multimodal model}
\label{sec:27bmm}

In addition to the MedGemma 4B multimodal and 27B \mbox{text-only} variants described earlier in this report, we are also releasing a MedGemma~27B~multimodal variant. The training methodology for this variant was the same as for the MedGemma 4B multimodal model with the addition of two training datasets: EHRQA (details in Section~\ref{sec:ehr}), to improve the model's inherent EHR understanding, and Chest ImaGenome \cite{wu2021chest,goldberger2000physiobank}, to enable anatomy localization on chest X-ray images.

The MedGemma 27B multimodal variant provides all of the capabilities of the 4B multimodal variant but with significantly improved language capabilities as well as improved EHR understanding and anatomy localization. Formal evaluation of this variant is ongoing but preliminary results can be found in Table~\ref{tab:27bmm}.

\begin{table}[thbp]
\centering
\footnotesize
\caption{\small\textbf{Preliminary results for MedGemma 27B multimodal model}}
\label{tab:27bmm}
\resizebox{\textwidth}{!}{
\begin{tabular}{@{}l l c c c c@{}}
\toprule
Task & Metric & \makecell[c]{MedGemma 4B \\ Multimodal} & \makecell[c]{MedGemma 27B \\ Text-only} & \makecell[c]{MedGemma 27B \\ Multimodal}\\
\midrule
\multicolumn{5}{@{}l}{\textbf{Text evaluation}} \\
\midrule
MedQA (4-op) & Accuracy & 64.4 & 87.7  & 85.3 \\
MedMCQA & Accuracy & 55.7 & 74.2  & 70.2 \\
PubMedQA & Accuracy & 73.4 & 76.8  & 77.2 \\
MMLU Med & Accuracy & 70.0 & 87.0  & 86.2 \\
MedXpertQA (text only) & Accuracy & 14.2 & 25.7  & 23.7 \\
AfriMed-QA & Accuracy & 52.0 & 84.0  & 72.0 \\
\midrule
\multicolumn{5}{@{}l}{\textbf{Electronic health record information retrieval}} \\
\midrule
EHRQA & Accuracy & 67.6 & 86.3 & 90.5 \\
\midrule
\multicolumn{5}{@{}l}{\textbf{Medical image classification}} \\
\midrule
MIMIC CXR & Average F1 for 5 conditions & 88.9 & N/A & 90.0 \\
CheXpert CXR & Average F1 for 5 conditions & 48.1 & N/A & 49.9 \\
CXR14 & Average F1 for 3 conditions & 50.1 & N/A & 45.3 \\
DermMCQA & Accuracy & 71.8 & N/A & 71.7 \\
PathMCQA & Accuracy & 69.8 & N/A & 71.6 \\
Eyepacs & Accuracy & 64.9 & N/A & 75.3 \\
\midrule
\multicolumn{5}{@{}l}{\textbf{Visual question answering}} \\
\midrule
SlakeVQA & Tokenized F1 & 72.3 & N/A & 70.3 \\
VQA-Rad & Tokenized F1 & 49.9 & N/A & 46.7 \\
\midrule
\multicolumn{5}{@{}l}{\textbf{Knowledge and reasoning}} \\
\midrule
MedXpertQA (text + MM) & Accuracy & 18.8 & N/A & 26.8 \\
\midrule
\multicolumn{5}{@{}l}{\textbf{Report generation}} \\
\midrule
MIMIC CXR & Radgraph F1 & 29.5$\dagger$ & N/A & 27.0$\dagger$ \\
\bottomrule
\end{tabular}
}
\begin{tablenotes}
  \scriptsize
  \item $\dagger$ Results from the pretrained checkpoints.
\end{tablenotes}
\end{table}


\end{document}